\documentclass[english,letterpaper]{article}
\usepackage[T1]{fontenc}
\usepackage[latin9]{inputenc}
\usepackage{babel}
\usepackage{float}
\usepackage{booktabs}
\usepackage{textcomp}
\usepackage{amsthm}
\usepackage{amsmath,amsfonts,amsthm,bm} 
\usepackage{soul}
\usepackage{wrapfig}
\usepackage{xcolor}
\usepackage{caption} 
\usepackage{amssymb}
\usepackage[makeroom]{cancel}
\usepackage{xcolor}
\usepackage{xargs}[2008/03/08]
\usepackage[unicode=true,pdfusetitle,
 bookmarks=true,bookmarksnumbered=false,bookmarksopen=false,
 breaklinks=false,pdfborder={0 0 1},backref=false,colorlinks=false]
 {hyperref}
\usepackage[acronym,smallcaps,nowarn,section,nogroupskip,nonumberlist]{glossaries}
\makeatletter

\providecommand{\tabularnewline}{\\}
\floatstyle{ruled}
\newfloat{algorithm}{tbp}{loa}
\providecommand{\algorithmname}{Algorithm}
\floatname{algorithm}{\protect\algorithmname}

\theoremstyle{plain}

\theoremstyle{plain}

\usepackage{algolyx}

\PassOptionsToPackage{numbers, compress}{natbib}

\usepackage[final]{neurips_2020}

\usepackage{graphicx}
\usepackage[acronym,smallcaps,nowarn,section,nogroupskip,nonumberlist]{glossaries}
\usepackage[nameinlink,capitalise]{cleveref}
\usepackage{amsthm}
\usepackage{amssymb}
\usepackage{color}
\usepackage{enumitem}
\usepackage[normalem]{ulem}

\usepackage{algolyx}
\usepackage{subfig}
\usepackage[nameinlink,capitalise]{cleveref}

\newcommand{\vbeta}{{\boldsymbol{\beta}}}
\DeclareMathOperator*{\argmax}{arg\,max} 

\DeclareMathOperator{\ELBO}{\textsc{elbo}}

\DeclareMathOperator{\TVO}{\textsc{tvo}}

\DeclareMathOperator{\Cov}{\mathrm{Cov}}
\DeclareMathOperator{\Var}{\mathrm{Var}}

\definecolor{seaborn_blue}{rgb}{0.12156862745098039, 0.4666666666666667, 0.7058823529411765}
\definecolor{seaborn_orange}{rgb}{1.0, 0.4980392156862745, 0.054901960784313725}
\definecolor{seaborn_green}{rgb}{0.17254901960784313, 0.6274509803921569, 0.17254901960784313}

\DeclareMathOperator{\vx}{\mathbf{x}}
\DeclareMathOperator{\vz}{\mathbf{z}}

\Crefname{algocf}{Algorithm}{Algorithms}
\crefname{algorithm}{Algorithm}{Algorithms}
\crefname{equation}{Equation}{Equations}
\crefname{figure}{Figure}{Figure}
\crefname{section}{§}{§§}
\Crefname{section}{§}{§§}
\crefformat{equation}{(#2#1#3)}

\newtheorem{theorem}{Theorem}

\makeatother

\providecommand{\lemmaname}{Lemma}
\providecommand{\theoremname}{Theorem}

\usepackage{multirow}
\usepackage{makecell}

\title{Formatting Instructions For NeurIPS 2020}

\author{
Vu Nguyen  \\University of Oxford \\  \texttt{vu@robots.ox.ac.uk}
\And
Vaden Masrani  \\ University of British Columbia \\  \texttt{vadmas@cs.ubc.ca} \And
  Rob Brekelmans\\ USC Information Sciences Institute \\  \texttt{brekelma@usc.edu}
  \And
 Michael A. Osborne \\University of Oxford \\  \texttt{mosb@robots.ox.ac.uk} \And 
  Frank Wood \\University of British Columbia\\  \texttt{fwood@cs.ubc.ca}
 }

\begin{document}

\global\long\def\se{\hat{\text{se}}}%

\global\long\def\interior{\text{int}}%

\global\long\def\boundary{\text{bd}}%

\global\long\def\new{\text{*}}%

\global\long\def\stir{\text{Stirl}}%

\global\long\def\dist{d}%

\global\long\def\HX{\entro\left(X\right)}%

\global\long\def\entropyX{\HX}%

\global\long\def\HY{\entro\left(Y\right)}%

\global\long\def\entropyY{\HY}%

\global\long\def\HXY{\entro\left(X,Y\right)}%

\global\long\def\entropyXY{\HXY}%

\global\long\def\mutualXY{\mutual\left(X;Y\right)}%

\global\long\def\mutinfoXY{\mutualXY}%

\global\long\def\xnew{y}%

\global\long\def\bx{\mathbf{x}}%

\global\long\def\bz{\mathbf{z}}%

\global\long\def\bu{\mathbf{u}}%

\global\long\def\bs{\boldsymbol{s}}%

\global\long\def\bk{\mathbf{k}}%

\global\long\def\bX{\mathbf{X}}%

\global\long\def\tbx{\tilde{\bx}}%

\global\long\def\by{\mathbf{y}}%

\global\long\def\bY{\mathbf{Y}}%

\global\long\def\bZ{\boldsymbol{Z}}%

\global\long\def\bU{\boldsymbol{U}}%

\global\long\def\bv{\boldsymbol{v}}%

\global\long\def\bn{\boldsymbol{n}}%

\global\long\def\bV{\boldsymbol{V}}%

\global\long\def\bK{\boldsymbol{K}}%

\global\long\def\bw{\vt w}%

\global\long\def\bbeta{\bm{\beta}}%

\global\long\def\bmu{\gvt{\mu}}%

\global\long\def\btheta{\boldsymbol{\theta}}%

\global\long\def\blambda{\boldsymbol{\lambda}}%

\global\long\def\bgamma{\boldsymbol{\gamma}}%

\global\long\def\bpsi{\boldsymbol{\psi}}%

\global\long\def\bphi{\boldsymbol{\phi}}%

\global\long\def\bpi{\boldsymbol{\pi}}%

\global\long\def\eeta{\boldsymbol{\eta}}%

\global\long\def\bomega{\boldsymbol{\omega}}%

\global\long\def\bepsilon{\boldsymbol{\epsilon}}%

\global\long\def\btau{\boldsymbol{\tau}}%

\global\long\def\bSigma{\gvt{\Sigma}}%

\global\long\def\realset{\mathbb{R}}%

\global\long\def\realn{\realset^{n}}%

\global\long\def\integerset{\mathbb{Z}}%

\global\long\def\natset{\integerset}%

\global\long\def\integer{\integerset}%

\global\long\def\natn{\natset^{n}}%

\global\long\def\rational{\mathbb{Q}}%

\global\long\def\rationaln{\rational^{n}}%

\global\long\def\complexset{\mathbb{C}}%

\global\long\def\comp{\complexset}%

\global\long\def\compl#1{#1^{\text{c}}}%

\global\long\def\and{\cap}%

\global\long\def\compn{\comp^{n}}%

\global\long\def\comb#1#2{\left({#1\atop #2}\right) }%

\global\long\def\nchoosek#1#2{\left({#1\atop #2}\right)}%

\global\long\def\param{\vt w}%

\global\long\def\Param{\Theta}%

\global\long\def\meanparam{\gvt{\mu}}%

\global\long\def\meanmap{\mathbf{m}}%

\global\long\def\logpart{A}%

\global\long\def\simplex{\Delta}%

\global\long\def\simplexn{\simplex^{n}}%

\global\long\def\dirproc{\text{DP}}%

\global\long\def\ggproc{\text{GG}}%

\global\long\def\DP{\text{DP}}%

\global\long\def\ndp{\text{nDP}}%

\global\long\def\hdp{\text{HDP}}%

\global\long\def\gempdf{\text{GEM}}%

\global\long\def\ei{\text{EI}}%

\global\long\def\rfs{\text{RFS}}%

\global\long\def\bernrfs{\text{BernoulliRFS}}%

\global\long\def\poissrfs{\text{PoissonRFS}}%

\global\long\def\grad{\gradient}%

\global\long\def\gradient{\nabla}%

\global\long\def\cpr#1#2{\Pr\left(#1\ |\ #2\right)}%

\global\long\def\var{\text{Var}}%

\global\long\def\Var#1{\text{Var}\left[#1\right]}%

\global\long\def\cov{\text{Cov}}%

\global\long\def\Cov#1{\cov\left[ #1 \right]}%

\global\long\def\COV#1#2{\underset{#2}{\cov}\left[ #1 \right]}%

\global\long\def\corr{\text{Corr}}%

\global\long\def\sst{\text{T}}%

\global\long\def\SST{\sst}%

\global\long\def\ess{\mathbb{E}}%

\global\long\def\Ess#1{\ess\left[#1\right]}%

\global\long\def\fisher{\mathcal{F}}%

\global\long\def\bfield{\mathcal{B}}%

\global\long\def\borel{\mathcal{B}}%

\global\long\def\bernpdf{\text{Bernoulli}}%

\global\long\def\betapdf{\text{Beta}}%

\global\long\def\dirpdf{\text{Dir}}%

\global\long\def\gammapdf{\text{Gamma}}%

\global\long\def\gaussden#1#2{\text{Normal}\left(#1, #2 \right) }%

\global\long\def\gauss{\mathbf{N}}%

\global\long\def\gausspdf#1#2#3{\text{Normal}\left( #1 \lcabra{#2, #3}\right) }%

\global\long\def\multpdf{\text{Mult}}%

\global\long\def\poiss{\text{Pois}}%

\global\long\def\poissonpdf{\text{Poisson}}%

\global\long\def\pgpdf{\text{PG}}%

\global\long\def\iwshpdf{\text{InvWish}}%

\global\long\def\nwpdf{\text{NW}}%

\global\long\def\niwpdf{\text{NIW}}%

\global\long\def\studentpdf{\text{Student}}%

\global\long\def\unipdf{\text{Uni}}%

\global\long\def\transp#1{\transpose{#1}}%

\global\long\def\transpose#1{#1^{\mathsf{T}}}%

\global\long\def\mgt{\succ}%

\global\long\def\mge{\succeq}%

\global\long\def\idenmat{\mathbf{I}}%

\global\long\def\trace{\mathrm{tr}}%

\glsdisablehyper{}
\newacronym{AIS}{ais}{Annealed Importance Sampling}
\newacronym{BQ}{bq}{Bayesian Quadrature}
\newacronym{AUC}{auc}{area under the curve}
\newacronym{ELBO}{elbo}{Evidence Lower Bound}
\newacronym{EUBO}{eubo}{evidence upper bound}
\newacronym{IS}{is}{importance sampling}
\newacronym{IWAE}{iwae}{Importance Weighted Autoencoder}
\newacronym{KL}{kl}{Kullback-Leibler}
\newacronym{RWS}{rws}{reweighted wake-sleep}
\newacronym{SGD}{sgd}{stochastic gradient descent}
\newacronym{SNIS}{snis}{self-normalized importance sampling}
\newacronym{TI}{ti}{Thermodynamic Integration}
\newacronym{TVI}{tvi}{thermodynamic variational inference}
\newacronym{TVO}{tvo}{Thermodynamic Variational Objective}
\newacronym{VAE}{vae}{Variational Autoencoder}
\newacronym{VI}{vi}{variational inference}
\newacronym{VIMCO}{vimco}{variational inference for Monte Carlo objectives}
\newacronym{WS}{ws}{wake-sleep}

\title{Gaussian Process Bandit Optimization of the Thermodynamic Variational Objective}

\maketitle


\begin{abstract}


Achieving the full promise of the Thermodynamic Variational Objective (TVO), a recently proposed variational lower bound on the log evidence involving a one-dimensional Riemann integral approximation, requires choosing a ``schedule'' of sorted discretization points.  This paper introduces a bespoke Gaussian process bandit optimization method for automatically choosing these points.  Our approach not only automates their one-time selection, but also dynamically adapts their positions over the course of optimization, leading to improved model learning and inference.  We provide theoretical guarantees that our bandit optimization converges to the regret-minimizing choice of integration points.   Empirical validation of our algorithm is provided in terms of improved learning and inference in Variational Autoencoders and Sigmoid Belief Networks.
\end{abstract}

%

\glsresetall

\section{Introduction}
The \gls{VAE} framework has formed the basis for a number of recent advances in unsupervised representation learning \cite{kingma2013vae, rezende2014vae,tschannen2018recent}.   Assuming a generative model involving latent variables, \gls{VAE}s perform maximum likelihood parameter estimation by optimizing the tractable \gls{ELBO} on the logarithm of the model evidence.   In doing so, the \gls{VAE} framework introduces an inference network, which seeks to approximate the true posterior over latent variables.  While the \gls{ELBO} is a common choice of variational inference objective, recent work has sought to improve the model learning \cite{burda2015importance, laddervae,nowozin2018debiasing, luo2020sumo} or inference aspects \cite{rezende2015variational,kingma2016improved,cremer2018inference, he2019lagging} of this task.
%

\begin{figure}
\centering
\includegraphics[width=.60\textwidth]{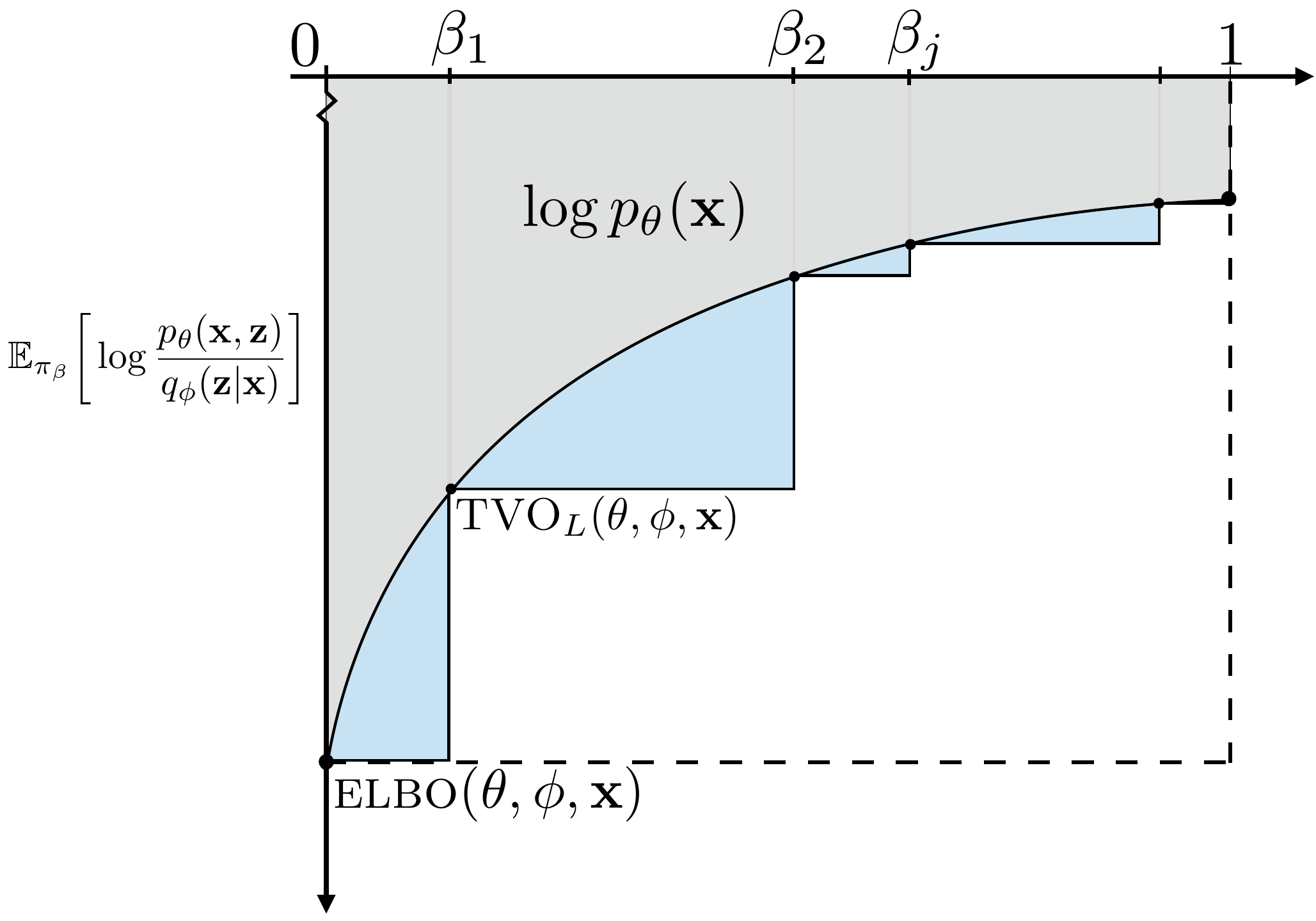}
\caption{The \gls{TVO} objective frames log likelihood estimation as a Riemann sum approximation to a 1-d integral, with the \gls{ELBO} as a special case for a single $\beta_0 = 0$.  The \gls{TVO} (area in blue) bounds the integral more tightly than the \gls{ELBO} (area within dotted lines).} \label{fig:tvo}
\end{figure}

In this work, we build upon the recent \gls{TVO}, which frames log-likelihood estimation as a one-dimensional integral over the unit interval \cite{masrani2019thermodynamic}.  
The integral is estimated using a Riemann sum approximation, as visualized in Figure \ref{fig:tvo}, yielding a natural family of variational inference objectives which generalize and tighten the \gls{ELBO}.

The choice of a $d$-dimensional vector of points $\vbeta = [\beta_0, \beta_1, ..., \beta_{d-1}]^T$ at which to construct this numerical approximation is an important hyperparameter for the \gls{TVO}, which we refer to as an ``integration schedule'' throughout this work.   Previous work \cite{masrani2019thermodynamic} uses a static integration schedule, and requires grid search over the choice of initial $\beta_1$.  However, since the shape of the integrand reflects the quality of the inference network (\S\ref{sec:preliminaries}), recent work \cite{brekelmans2019tvo} suggests that this scheduling procedure may be improved by dynamically choosing  $\vbeta$ over the course of training.  Our proposed approach also allows the \gls{TVO} to be adapted to different model architectures and schedule dimensionality without the need for grid search. 

Our primary contribution is to automate the choice of integration schedules using a Gaussian process bandit optimization.  We first demonstrate that maximizing the \gls{TVO} objective as a function of $\vbeta$ is equivalent to a regret-minimization problem, where the black-box reward function reflects improvement in the objective for a given choice of schedule.  We model this reward function over the course of training epochs using a time-varying Gaussian process (GP).  Our entire procedure amounts to 1) choosing $\vbeta$ to maximize an acquisition function in our surrogate GP model, 2) observing the reward function as the improvement in the \gls{TVO} objective over one or more epochs of training with the chosen schedule, and 3) using these observations to update the GP model and select a new $\vbeta$. 
Our bandit algorithm is optimal in the sense of converging to a global regret-minimizing solution, {as in the time-varying GP bandit optimization approach \cite{bogunovic2016time}.}
By choosing $\vbeta$ to maximize an acquisition function that balances exploration and performance, our algorithm achieves global guarantees despite the non-convexity of the reward function.  Further, our approach is directly aligned with the goal of improved model learning and inference, as the bandit reward function tracks the variational objective over the course of training. 

We review the \gls{TVO} framework in \S \ref{sec:preliminaries}, before presenting our bandit optimization approach in \S \ref{sec:bandits}.  We provide details of our time-varying Gaussian process model and discuss its convergence properties in \S \ref{sec:gp}.  Finally, we demonstrate that our method can improve both model learning and inference in Variational Autoencoders and Sigmoid Belief Networks,  in \S \ref{sec:experiments}.

\section{The Thermodynamic Variational Objective (TVO)}
\label{sec:preliminaries}

Assuming a generative model $p_{\theta}(\vx,\vz)$, we are interested in maximizing the log-likelihood $\log p_{\theta}(\vx) = \log \int p_{\theta}(\vx,\vz) d\vz $ over parameters $\theta$, given the empirical data $\vx$.   However, this is intractable due to the integral over the latent variables $\vz$.  Variational inference methods \cite{blei2017variational} often seek to maximize the tractable \gls{ELBO} instead, obtained by introducing an approximate posterior $q_{\phi}(\vz|\vx)$ and optimizing the objective
\begin{align}
    \ELBO(\theta, \phi, \vx) = \log p_{\theta}(\vx) - D_{\text{KL}}[q_{\phi}(\vz|\vx) \, || \, p_{\theta}(\vz|\vx)] = \mathbb{E}_{q_{\phi}(\vz|\vx)} \left[\log \frac{p_{\theta}(\vx,\vz)}{q_{\phi}(\vz|\vx)} \right] \, .  \label{eq:elbo}
\end{align}
\gls{TI} \cite{ogata1989monte,frenkel2001understanding,gelman1998simulating} is a common technique for estimating (ratios of) partition functions in statistical physics, which  
instead frames estimating $\log p_{\theta} (\vx)$ as a one-dimensional integral over a geometric mixture curve parameterized by $\beta$.\footnote{Here $\beta$ is a scalar to be consistent with notation in \cite{masrani2019thermodynamic} In the remainder of the paper, we let $\vbeta$ hold the sorted vector of discretization points $\vbeta = [\beta_0, \beta_1, ..., \beta_{d-1}]^T$, so that each $\beta_j$ specifies a $\pi_{\beta_j}(\vz|\vx)$ in \eqref{eq:exp_family}. } In particular, for the \gls{TVO}, this curve interpolates between the approximate posterior $q_{\phi}(\vz|\vx)$ and true posterior $p_{\theta}(\vz|\vx)$.  Following \cite{brekelmans2019tvo} this mixture curve can be interpreted as an exponential family of distributions over $\vz$ given $\vx$
\begin{align}
    \pi_{\beta}(\vz|\vx) &= q_{\phi}(\vz|\vx) \exp \{ \beta \cdot \log \frac{p_{\theta}(\vx, \vz)}{q_{\phi}(\vz|\vx)} - \log Z_{\beta}(\vx) \} \label{eq:exp_family}\\
    \text{where } Z_{\beta}(\vx) &= \int q_{\phi}(\vz|\vx)^{1-\beta} p_{\theta}(\vx, \vz)^{\beta}  d\vz \, . \nonumber
\end{align}
Noting that $\log Z_{0}(\vx) = 0$ and $\log Z_{1}(\vx) = \log p_{\theta} (\vx)$, \gls{TI} now applies the fundamental theorem of calculus to write the model evidence as an integral
\begin{align}
\log p_{\theta} (\vx) - 0 &= \int_0^1 \frac{\partial}{\partial \beta} \log Z_{\beta}(\vx) d\beta =\int_0^1 \mathbb{E}_{\pi_{\beta}}\left[\log \frac{p_{\theta}(\vx, \vz)}{q_{\phi}(\vz|\vx)} \right] d\beta, \label{eq:ti_integral}
\end{align}
where we have used the known property of exponential families \cite{Wainwright_Jordan_08graphical} that the derivative of $\log Z_{\beta}(\vx)$ with respect to $\beta$ matches the expected sufficient statistics \cite{masrani2019thermodynamic, brekelmans2019tvo}.
\citet{masrani2019thermodynamic} use \gls{SNIS} to estimate each term in the integrand, with $S$ importance samples and $q_{\phi}(\vz|\vx)$ as the proposal for each $\beta$ 
\begin{align}
   \mathbb{E}_{\pi_{\beta}} [ \cdot ] \approx \sum \limits_{\ell = 1}^{S} \frac{w_{\ell}^{\beta}}{  \sum_{\ell} w_{\ell}^{\beta}} [ \cdot ], \quad w_{\ell} = \frac{p_{\theta}(\vx, \vz_{\ell})}{q_{\phi}(\vz_{\ell}|\vx)}, \, \, \vz_{\ell} \sim q_{\phi}(\vz|\vx). \label{eq:snis}
\end{align}


Since $\log Z_{\beta}(\vx)$ is convex \cite{Wainwright_Jordan_08graphical}, we know that the integrand in \cref{eq:ti_integral} is an increasing function of $\beta$.  Thus, we can obtain lower and upper bounds using left- and right-Riemann sums, respectively, over a discrete partition 
$\vbeta$ of the unit interval.  The left-Riemann sum then defines the \gls{TVO} lower bound 
\begin{align}
    \TVO(\theta, \phi, \vbeta, \vx) := \sum \limits_{j=0}^{d-1} (\beta_{j+1} - \beta_j) \, \mathbb{E}_{\pi_{\beta_j}} \left[ \log \frac{p_{\theta}(\vx, \vz)}{q_{\phi}(\vz|\vx)} \right],
    \label{eq:left_Riemann_sum}
\end{align}
where $\vbeta = [\beta_j]_{j=0}^{d-1}$ with $\beta_0=0$ and $\beta_j < \beta_{j+1}$.  Note that the single-term left-Riemann sum with $\vbeta = \beta_0 = 0$ matches the \gls{ELBO} in \eqref{eq:elbo}, since $\pi_0(\vz|\vx) = q_{\phi}(\vz|\vx)$.  However, how to choose intermediate $ \beta_j$ for $d>1$ remains an interesting question, which we proceed to frame as a bandit problem. 

\section{From Evidence Maximization to Regret Minimization}\label{sec:bandits}

We view the vector $\vbeta \in [0,1]^d$ as an arm \cite{Auer2002nonstochastic} to be pulled in a continuous space, given a fixed resource of $T$ training epochs. After each round, we receive an estimate of the log evidence $\mathcal{L}$,  from which we will construct a reward function. An important feature of our problem is that the integrand in \cref{fig:tvo} changes between rounds as training progresses. Thus, our multi-armed bandit problem is said to be \textit{time-varying}, in that the optimal arm and reward function depend on round $t$.

More formally, we define the time-varying reward function $f_{t}:[0,1]^d\rightarrow\mathbb{R}$ which takes an input $\vbeta_{t}$ and produces reward $f_{t}(\vbeta_{t})$. At each round we get access to a noisy reward ${y_t=f_{t}(\vbeta_{t})+\epsilon_t}$ where we assume Gaussian noise $\epsilon_{t}\sim\mathcal{N}(0,\sigma_{f}^{2})$.  We aim to maximize the cumulative reward $\sum_{t=1}^{T/w}f_{t}(\vbeta_{t})$ across $T/w$ rounds, where $w$ is a divisor of $T$ and will later control the ratio of bandit rounds $t$ to training epochs $i$.


Maximizing the cumulative reward is equivalent to minimizing the \textit{cumulative regret}
\begin{align}
    R_{T/w} := \sum_{t=1}^{T/w} f_{t}(\vbeta_t^*) - f_{t}(\vbeta_t), \label{eq:cumulative_regret}
\end{align}
where $r_t := f_{t}(\vbeta_t^*) - f_{t}(\vbeta_t)$ is the \textit{instantaneous regret} defined by the difference between the received reward $f_{t}(\vbeta_t)$ and maximum reward attainable $ f_{t}(\vbeta_t^*)$ at round $t$. The regret, which is non-negative and monotonic, is more convenient to work with than the cumulative reward and will allow us to derive upper bounds in \S \ref{sec:convergence}.


In order to translate the problem of maximizing the log evidence as a function of $\vbeta$ into the bandits framework, we define a time-varying reward function $f_t(\vbeta_t)$.   We construct this reward in such a way that minimizing the cumulative regret is equivalent to maximizing the final log evidence estimate $\mathcal{L}_T := \log p_{\theta_{T}}(\vx)$, i.e., such that $\min R_{T/w} = \max \mathcal{L}_T$.
%

Such a reward function can be defined by partitioning the $T$ training epochs into windows of equal length $w$, and defining the reward for each window $t \in \{0, 1, ..., T/w - 1\}$ 
\begin{align}
    f_{t}(\vbeta_{t}):=\mathcal{L}_{w(t+1)} - \mathcal{L}_{wt}\label{eq:reward_function}
\end{align}
as the difference between the \gls{TVO} log evidence estimate one window-length in the future $\mathcal{L}_{w(t+1)}$ and the present estimate $\mathcal{L}_{wt}$. Then, the cumulative reward is given by a telescoping sum over windows 
\begin{align}
\sum_{t=0}^{T/w~-~1}f_{t}(\vbeta_{t}) &= 
\big(\mathcal{L}_{w} - \mathcal{L}_{0}\big) + 
\big(\mathcal{L}_{2w} - \mathcal{L}_{w}\big) + ... + \big(\mathcal{L}_{(T/w)w} - \mathcal{L}_{(T/w~-~1)w}\big)\\
                                  &= \mathcal{L}_T - \mathcal{L}_0, \label{eq:cumulative_reward_function}
\end{align}
where $\mathcal{L}_0$ is the initial (i.e. untrained) loss. Recalling the definition of cumulative regret in Eq. \cref{eq:cumulative_regret},
\begin{align} \label{eq:equivalent_minregret_maxTVO}
    \min R_{T/w} &= \min\left(\sum_{t=0}^{T/w~-~1}f_{t}(\vbeta_t^*)\right) - \left(\sum_{t=0}^{T/w~-~1}f_{t}(\vbeta_t)\right)\\
                 &= \min\biggl(\left(\mathcal{L}^*_T - \mathcal{L}_0\right) - \left(\mathcal{L}_T - \mathcal{L}_0\right)\biggr)  \label{eq:L_T_optimal}\\
                 &= \min~(\text{const} - \mathcal{L}_T) \\
                 &= \max\mathcal{L}_T.
\end{align}
Therefore minimizing the cumulative regret for the reward function defined by Eq. \cref{eq:reward_function} is equivalent to maximizing the log evidence on the final epoch. Next, we describe how to design an optimal decision mechanism to minimize the cumulative regret $R_{T/w}$ using Gaussian processes.


\begin{algorithm}
\caption{GP-bandit for TVO (high level)\label{alg:TVO_GPTV_highlevel}}
\begin{algor}
\item [{{*}}] Input: schedule dimension $d$, reward function $f_{t}(\vbeta_{t})$ where $\vbeta_t \in [0,1]^d$ , update frequency $w$
\end{algor}
\begin{algor}[1]
\item [{for}] $t=1....T$
\item [{{*}}] Train $\theta, \phi$ for one epoch using TVO and previously $\vbeta_{t-1}$, evaluate $\mathcal{L}_{t}$ from Eq. (\ref{eq:left_Riemann_sum})
\item [{if}] mod$(t,w)=0$ : time to update $\vbeta_t$
\item [{{*}}] Estimate the utility $y_{t}=\mathcal{L}_{t}-\mathcal{L}_{t-w}$ and augment
$\left(\vbeta_{t-1},t,y_{t}\right)$ into training data $D_t$
\item [{{*}}] Fit a time-varying, permutation invariant GP to $D_{t}$ 
\item [{{*}}] Estimate GP predictive mean $\mu_{t}(\vbeta)$ and uncertainty $\sigma_{t}(\vbeta)$ from Eqs.
(\ref{eq:GP_mean},\ref{eq:GP_var}) 
\item [{{*}}] Select $\vbeta_{t}=\arg\max \mu_{t}(\vbeta)+\sqrt{\kappa_{t}}\sigma_{t}(\vbeta)$ where $\kappa_{t}$ is from Theorem \ref{thm:theorem_sublinear_regret}
\item [{endif}]~
\item [{endfor}]~
\end{algor}
\end{algorithm}
\section{Minimizing Regret with Gaussian Processes}\label{sec:gp}
There are two unresolved problems with the reward function defined in Eq. (\ref{eq:reward_function}) which still must be addressed. The first is that it is not in fact computable, due to its use of future observations. The second is that it ignores the ordering constraint required for $\vbeta$ to be a valid Riemann partition.

We can handle both by problems by using a permutation-invariant Gaussian process to form a surrogate for the reward function $f_{t}(\vbeta_{t})$. The surrogate model will be updated by past rewards, and used in place of $f_{t}(\vbeta_{t})$ to select the next schedule at the current round, as described in Algorithm \ref{alg:TVO_GPTV_highlevel}.

In \S \ref{sec:TVGP} we formally define how to use (time-varying) Gaussian processes in bandit optimization, before describing how our permutation-invariant kernel can be used to solve the problem of ordering constraints on $\vbeta$ in \S \ref{sec:perm_invar}. Finally in \S \ref{sec:convergence} we provide a theoretical guarantee that our bandit optimization will converge to the regret-minimizing choice of  $\vbeta$. 

\subsection{Time-varying Gaussian processes for Bandit Optimization} \label{sec:TVGP}
A popular design   in handling time-varying functions \cite{Krause_2011Contextual,Swersky_2013Multi,klein2017fast,nyikosa2018adaptive} such as $f_{t}(\vbeta_{t})$ is to jointly model the spatial and temporal dimensions using a product of covariance functions ${k=k_{\beta}\otimes k_{T}}$, where ${k_{\beta}:[0, 1]^d\times [0, 1]^d\rightarrow\mathbb{R}}$ is a spatial covariance function over actions, ${k_{T}:\mathbb{N}\times\mathbb{N}\rightarrow\mathbb{R}}$ is a temporal covariance function, and ${k:\mathbb{R}_+^{d+1}\times\mathbb{R}_+^{d+1}\rightarrow\mathbb{R}}$.

Under this joint modeling framework, the GP is defined as follows. At round $t$ we have the history of rewards $\bm{y}_t = [y_0, ..., y_t]^T$ and sample points $\bm{X}_t = \{\bm{x}_0, ..., \bm{x}_t\}$, where we define $\bm{x}_t \in \mathbb{R}^{d+1}$ to be the concatenation of $\vbeta_t$ and timestep $t$, i.e $\bm{x}_t:= [\vbeta_t, t]^T$. Then the time-varying reward function is GP-distributed according to
\begin{align}
    f_t \sim GP\bigl(0, k(\bm{x}, \bm{x}')\bigr) \quad \quad \quad
    \text{where } k(\bm{x}, \bm{x}') := k_{\beta}(\vbeta, \vbeta') \times k_{T}(t, t'),
\end{align}
where we have assumed zero prior mean for simplicity. For theoretical convenience we follow \cite{bogunovic2016time} and choose ${k_{T}(t,t')=(1-\omega)^{\frac{|t-t'|}{2}}}$, where $\omega$ is a ``forgetting-remembering'' trade-off parameter learned from data. We describe $k_{\beta}(\vbeta, \vbeta')$ in \S \ref{sec:perm_invar}.

Using standard Gaussian identities \cite{Bishop_2006pattern,Rasmussen_2006gaussian}, the posterior predictive is also GP distributed, with mean and variance given by
\begin{align}
\mu_t(\vbeta_*) &= \bm{k}_t(\vbeta_*)^T\left(\bm{K}_t + \sigma_{f}^{2}\bm{I} \right)^{-1}\bm{y}_t \label{eq:GP_mean} \\
\sigma_t^2(\vbeta_*) &= \bm{k}_t(\vbeta_*, \vbeta_*) - \bm{k}_t(\vbeta_*)^T\left(\bm{K}_t + \sigma_{f}^{2}\bm{I} \right)^{-1}\bm{k}_t(\vbeta_*)\label{eq:GP_var}
\end{align}
where $\bm{k}_t(\vbeta) = [k(\bm{x}_0, \bm{x}), ... , k(\bm{x}_t, \bm{x})]^T$ and $\bm{K}_t = [k(\bm{x}, \bm{x}')]_{\bm{x}, \bm{x}' \in \bm{X}_t}$. Using this permutation-invariant, time-varying GP we can select $\vbeta_{t+1}$ by maximizing a linear combination of the GP posterior mean and variance w.r.t $\vbeta_t$
\begin{align}
\vbeta_{t+1}&= \argmax_{\vbeta_t}\mu_t(\vbeta_t) + \sqrt{\kappa_{t}}\sigma_t(\vbeta_t),
\label{eq:acquisition_function}
\end{align}
where Eq. (\ref{eq:acquisition_function}) is referred to as an \textit{acquisition function} and $\kappa_{t}$ is its exploration-exploitation trade-off parameter. We note that there are other acquisition functions available \cite{Jones_1998Efficient,Hennig_2012Entropy,Hernandez_2014Predictive}. Our acquisition function, Eq. (\ref{eq:acquisition_function}), is the time-varying version of GP-UCB \cite{Srinivas_2010Gaussian,bogunovic2016time}, which allows us to obtain convergence results in \S\ref{sec:convergence} and set $\kappa_{t}$ in Theorem \ref{thm:theorem_sublinear_regret}. 


\subsection{Ordering Constraints and Permutation Invariance}\label{sec:perm_invar}



Recall that the vector $\vbeta=\left[\beta_{0},...,\beta_{d-1}\right]^T$ holds the locations of the left Riemann integral approximation in Eq. \eqref{eq:left_Riemann_sum}. In order for the left Riemann approximation to the \gls{TVO} to be sensible, there must be an ordering constraint imposed on $\bbeta$ such that $0 < \beta_1 < ... < \beta_{d-1} < 1$.  We model this in our GP using a projection operator $\Phi$ which imposes the constraint by sorting the vector $\vbeta$.  Applying $\Phi$ within the spatial kernel, we obtain
\begin{align}
    k_\beta\bigl(\bbeta, \bbeta' \bigr):=k_{S} \bigl(\Phi(\bbeta),\Phi(\bbeta') \bigr) \, .
\end{align}  
This projection does not change the value of our acquisition function, and maintains the positive definite for any covariance function for the spatial $k_S$, e.g., Matern, Polynomial.
 We then optimize the acquisition function via a projected-gradient approach.  If a $\bbeta_t$ iterate leaves the feasible set after taking a gradient step, we project it back into the feasible set using $\Phi$ and continue. 
   We note that existing work in the GP literature has considered such projection operations in various contexts \cite{snoek2014input,van2018learning}. 

\subsection{Convergence Analysis} \label{sec:convergence}
In Eq.
(\ref{eq:equivalent_minregret_maxTVO}), we showed that maximizing the \gls{TVO} objective function $\mathcal{L_T}$ as a function of $\vbeta_t$ is equivalent to minimizing the cumulative regret $R_{T/w}$ by sequential optimization within the bandit framework. Here, the subscript $T/w$ refers to the number of bandit updates given the maximum epochs $T$ and the update frequency $w$ where $w \ll T$. 

We now derive an upper bound on the cumulative regret, and show that it asymptotically
goes to zero as $T$ increases, i.e., $\lim_{T\rightarrow\infty}\frac{R_{T/w}}{T}=0$.
Thus, our bandit will converge to choosing $\vbeta_{T}$ which yields the optimal value of the \gls{TVO} objective $\mathcal{L^*_T}$ for model parameters at step $T$.

We present the main theoretical result in Theorem \ref{thm:theorem_sublinear_regret}.  
Our \gls{TVO} framework mirrors the standard time-varying GP bandit optimization, and thus inherits convergence guarantees from \citet{bogunovic2016time}.  However, as discussed in Appendix \S\ref{app:theory}, we provide a tighter bound on the mutual information gain $\gamma_{T/w}$ which may be of wider interest. 





\begin{theorem} \label{thm:theorem_sublinear_regret}
Let the domain $\mathcal{D}\subset[0,1]^{d}$ be compact and convex.  Let $L_t \ge0$ be the Lipschitz constant for the reward function at time $t$.    Assume that the covariance function $k$ is almost surely continuously differentiable, with $f\sim GP(0,k)$.  Further, for $t\le T$ and $j\le d$, we assume
\begin{align*}
    \text{Pr} \left(\sup\left| \partial f_{t}(\vbeta_{t}) / \partial\vbeta_{t}^{(j)}\right|\ge L_{t}\right)\le ae^{-\left(L_{t}/b\right)^{2}} \quad 
\end{align*}
for appropriate choice of $a$ and $b$ corresponding to $L_t$.

For $\delta\in(0,1)$, we write  $\kappa_{T/w}=2\log\frac{\pi^{2}T^{2}}{2\delta w^2}+2d\log db{\frac{T^{2}}{w^2}} \sqrt{\log\frac{da\pi^{2}T^{2}}{2\delta w^2}}$
and $C_{1}=8/\log(1+\sigma_{f}^{2})$.  Then, after $T/w$ time steps, our algorithm satisfies 
\begin{align*}
R_{T/w}=\sum_{t=1}^{T/w}f_{t}(\vbeta_{t}^{*})-f_{t}(\vbeta_{t}) & \le\sqrt{\gamma_{T/w} \cdot C_{1} \cdot \kappa_{T/w} \cdot T/w}+2
\end{align*}
with probability at least $1-\delta$, where $\gamma_{T/w}$ is the maximum information gain for the time-varying covariance function (see below). 
\end{theorem}


In the above theorem, the quantity $\gamma_{T/w}$ measures the maximum information gain obtained about the reward function after pulling $T/w$ arms \cite{Srinivas_2010Gaussian, bogunovic2016time}. 
In the Appendix \S\ref{app:theory}, we show that $\gamma_{T/w} \le \left(1+ T / \left[w \tilde{N} \right]\right) \left(\gamma^{\vbeta}_{\tilde{N}}+\sigma_{f}^{-2}\tilde{N}^{5/2}\omega\right)$,
where $\tilde{N}\in\left\{ 1,...,T/w\right\}$ denotes a time-varying block length, and $\gamma^{\vbeta}_{\tilde{N}}$ is defined with respect to the covariance kernel for $\vbeta$. 
For our particular choice of exponentiated-quadratic kernel, the maximum information gain scales as $\gamma^{\vbeta}_{\tilde{N}} \le  \mathcal{O}( \log \tilde{N}^{d+1})$ \cite{Srinivas_2010Gaussian}. 
Compared with \cite{bogunovic2016time}, our proof tightens the upper bound on $\gamma_{T/w}$ from $\mathcal{O}(\tilde{N}^{3})$ to $\mathcal{O}(\tilde{N}^{5/2})$.

Combining these terms, we can then write the bound as $R_{T/w} \lesssim \mathcal{O}(\sqrt{\left( \left[ \log \tilde{N}^{d+1} + \sigma_{f}^{-2}\tilde{N}^{5/2}\omega\right] T/w \right)}$, which is sublinear in $T$ when the function $f$ becomes time-invariant, i.e.,  $\omega \rightarrow 0$. In contrast, the sublinear guarantee does not hold when the time-varying function is non-correlated, i.e., $\omega=1$, in which case the time covariance matrix becomes identity matrix. The bound is tighter for lower schedule dimension $d$.



\begin{figure*}
\includegraphics[width=0.33\columnwidth]{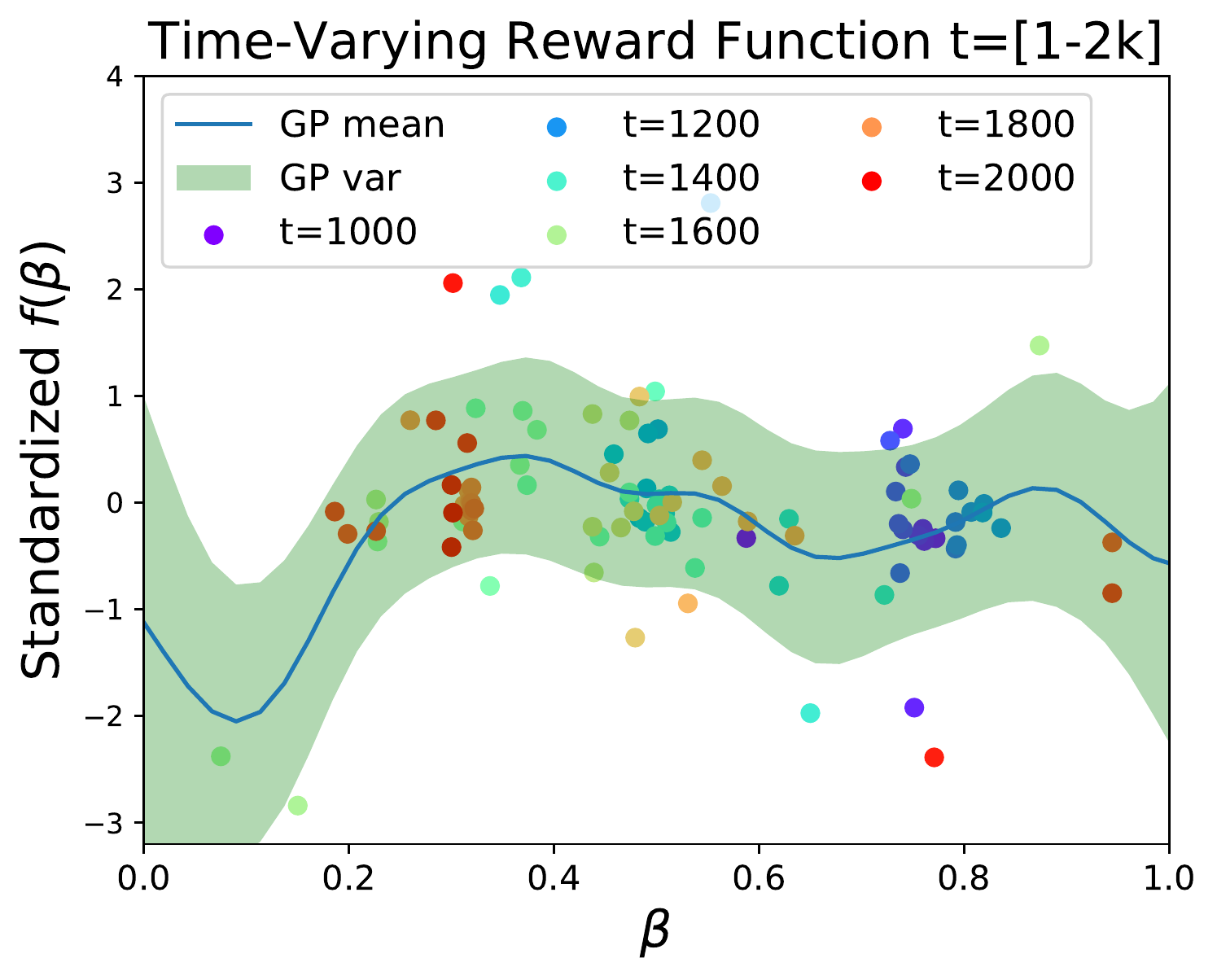}
\includegraphics[width=0.33\columnwidth]{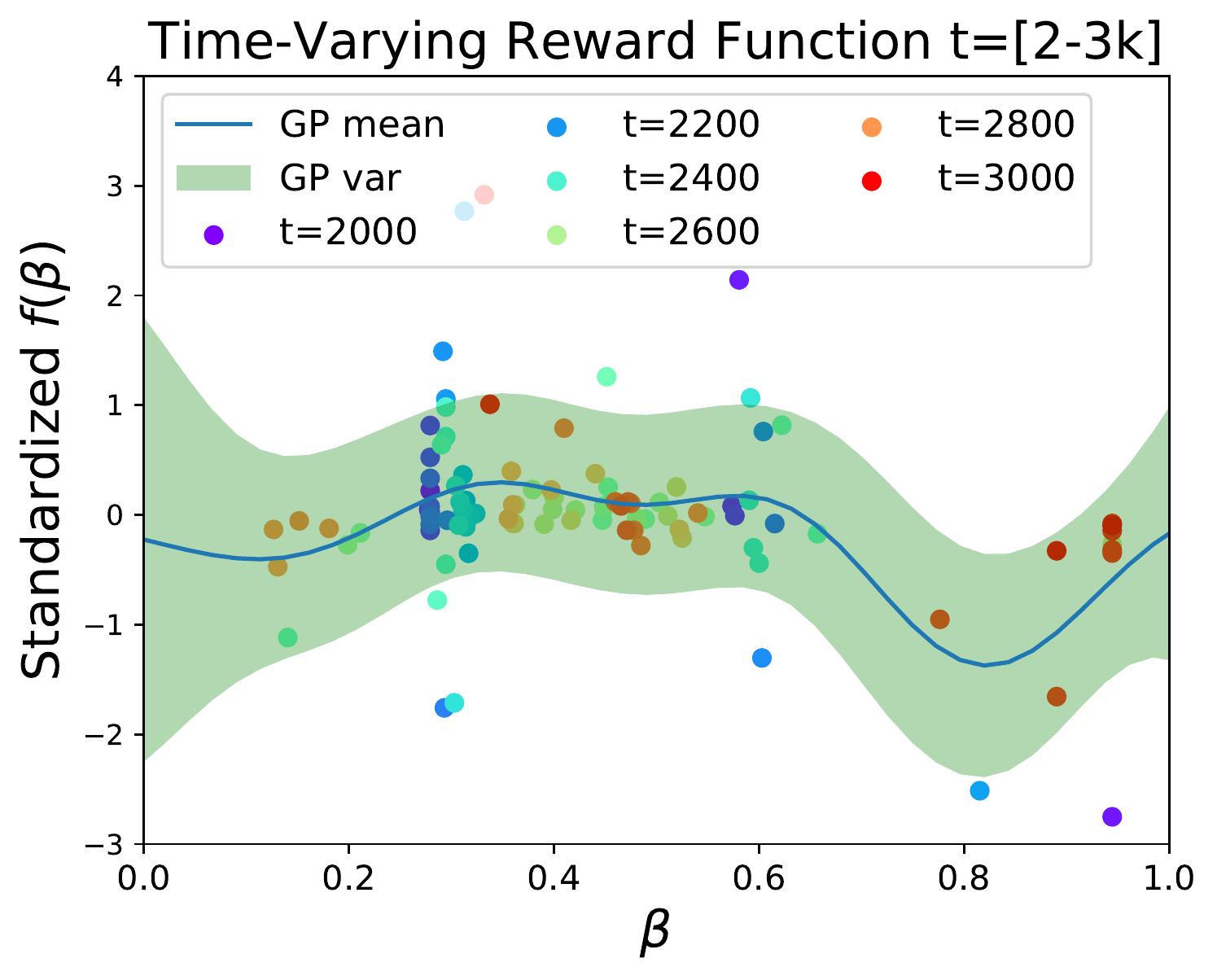}
\includegraphics[width=0.33\columnwidth]{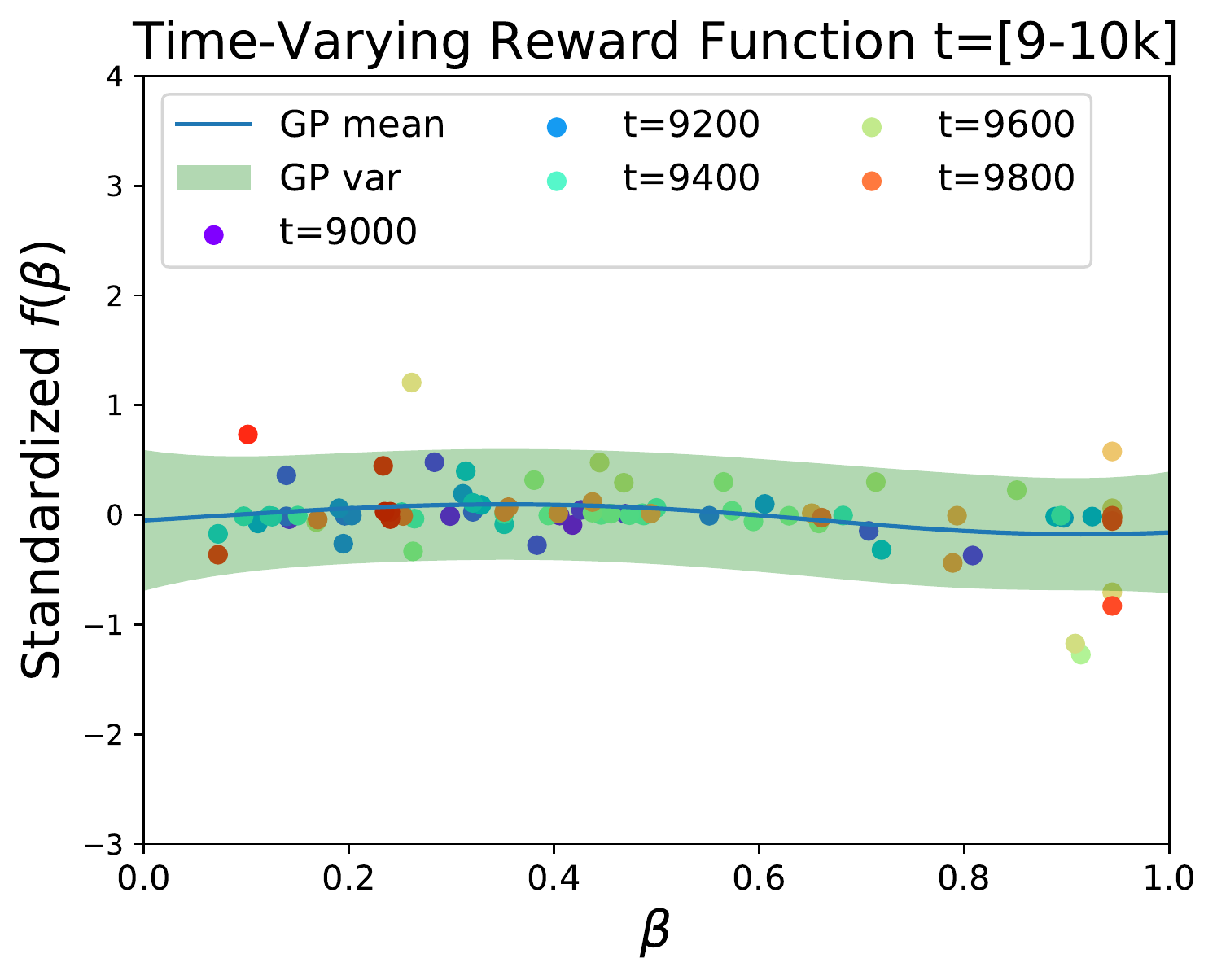}
\caption{Time-varying reward function $f_t(\bbeta_t)$ after 2k, 3k, and 10k training epochs, with bandit choices of a single intermediate $\beta_1$ (i.e. $\vbeta = [0, \beta_1]$) colored by timestep.  Scattered $\beta_1$ in neighboring epochs indicate `exploration', while similarly colored values of $\beta_1$ in regions where the GP mean, or predicted reward, is high indicate `exploitation'. 
} \label{fig:exmaple_time_reward}
\end{figure*}


\section{Experiments}\label{sec:experiments}

We demonstrate the effectiveness of our method for training \gls{VAE}s \cite{kingma2013vae} on MNIST and Fashion MNIST, and a Sigmoid Belief Network \cite{mnih2014neural} on binarized MNIST and binarized Omniglot, using the \gls{TVO} objective.  In \cref{app:experiments}, we explore learning and inference in a discrete probabilistic context-free grammar  \cite{le2019revisiting}, showing that the \gls{TVO} objective and our bandit optimization can translate to other learning settings. In addition, we run ablation studies using random choices of $\vbeta$ and a GP without permutation invariance, and compare the runtime and performance of our method with grid search.  Our code is available at \url{http://github.com/ntienvu/tvo_gp_bandit}.

\begin{wrapfigure}{r}{0.49\textwidth}
\centering
\includegraphics[width=0.45\columnwidth]{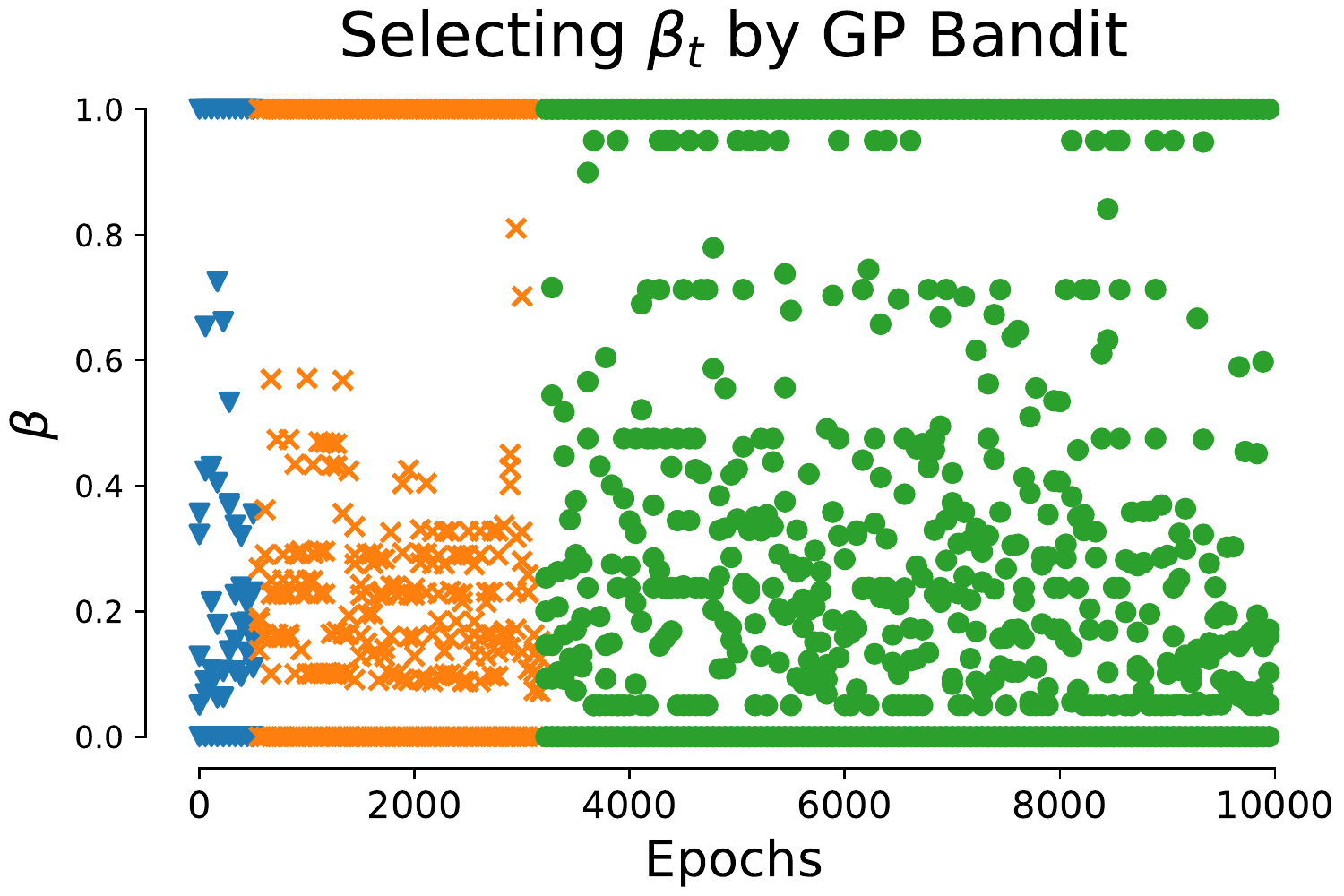}
\includegraphics[width=0.49\columnwidth]{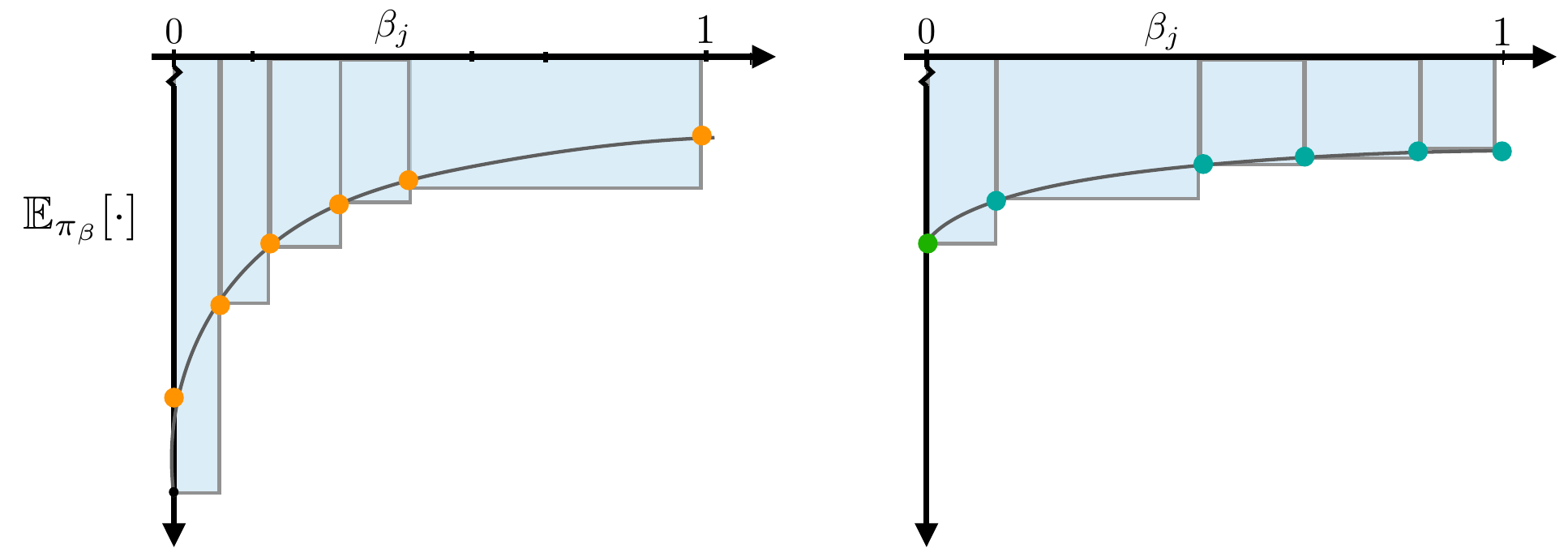}
\caption{Bandit-chosen $\vbeta$ over time on MNIST using $d=5$.  We can interpret the $\vbeta$ selection process in $3$ phases: (\textcolor{seaborn_blue}{blue}) random selection in initial epochs; (\textcolor{seaborn_orange}{orange}) focusing on small values $\beta_i < 0.5$ as training progresses;  (\textcolor{seaborn_green}{green}) moving toward $\beta_i=1$ as learning approaches convergence. The bottom panel illustrates a hypothetical integrand curve and $\vbeta$ selections at intermediate (left) and later (right) epochs. \\}\label{fig:beta_overtime}
\vspace{-25pt}
\end{wrapfigure}


\vspace*{-.1cm}
\paragraph{Experimental Setup:}
We evaluate our GP-bandit for  $S \in  \{10,50 \}$ and $d \in \{ 2, 5, 10, 15 \}$ and, for each configuration, train until convergence using $5$ random seeds.  Note that, for each setting of $d$, we implicitly include $\beta_0 = 0$ and append 1 to the vector $\vbeta$ to perform the integration in Eq. (\ref{eq:left_Riemann_sum}).

For each $S, d$ configuration, we compare against three baseline integration schedules: log-uniform spacing in the interval $[\beta_1, 1]$, linear-uniform spacing in the interval $[0,1]$, and the moments schedule of \cite{brekelmans2019tvo, grosse2013annealing}, which corresponds to uniform spacing along the y-axis.
For log/linear-uniform spacing, we set $\beta_1 = 0.025$ for all experiments, reflecting the results of grid search in \cite{masrani2019thermodynamic}. We use a fixed model architecture for all experiments, which we describe in Appendix \ref{app:ExperimentalSetup}.  

To obtain the bandit feedback in Eq. \eqref{eq:reward_function}, we use a fixed, linear schedule with $d=50$ for calculating $\mathcal{L}_t$ with Eq. \eqref{eq:left_Riemann_sum}.  This yields a tighter $\log p_{\theta}(\vx)$ bound, decouples reward function evaluation from model training and schedule selection in each round, and is still efficient using \gls{SNIS} in Eq. \eqref{eq:snis}.  We limit the value of $d$ for \gls{TVO} training following observations of deteriorating performance in \cite{masrani2019thermodynamic}.
\paragraph{GP Implementation:}
For GP modeling, we use an exponentiated quadratic covariance function for $k_{\beta}$ and estimate hyperparameters via type II maximum likelihood estimation~\cite{Rasmussen_2006gaussian}. We use multi-start BFGS \cite{fletcher2013practical} to optimize the acquisition function in Eq. (\ref{eq:acquisition_function}). We set the update frequency $w=6$ initially and increment $w$ by one after every $10$ bandit iterations to account for smaller objective changes later in training, and update early if $\mathcal{L}_t \le -0.05$.  We found that selecting $\beta_j$ too close to either $0$ or $1$ could negatively affect performance, and thus restrict  $\vbeta \in [0.05,0.95]^d$ in all experiments. We follow a common practice to standardize with the running average the utility score $y \sim \mathcal{N}(0,1)$ for robustness.





\begin{figure*}[t]

\includegraphics[width=0.5\columnwidth]{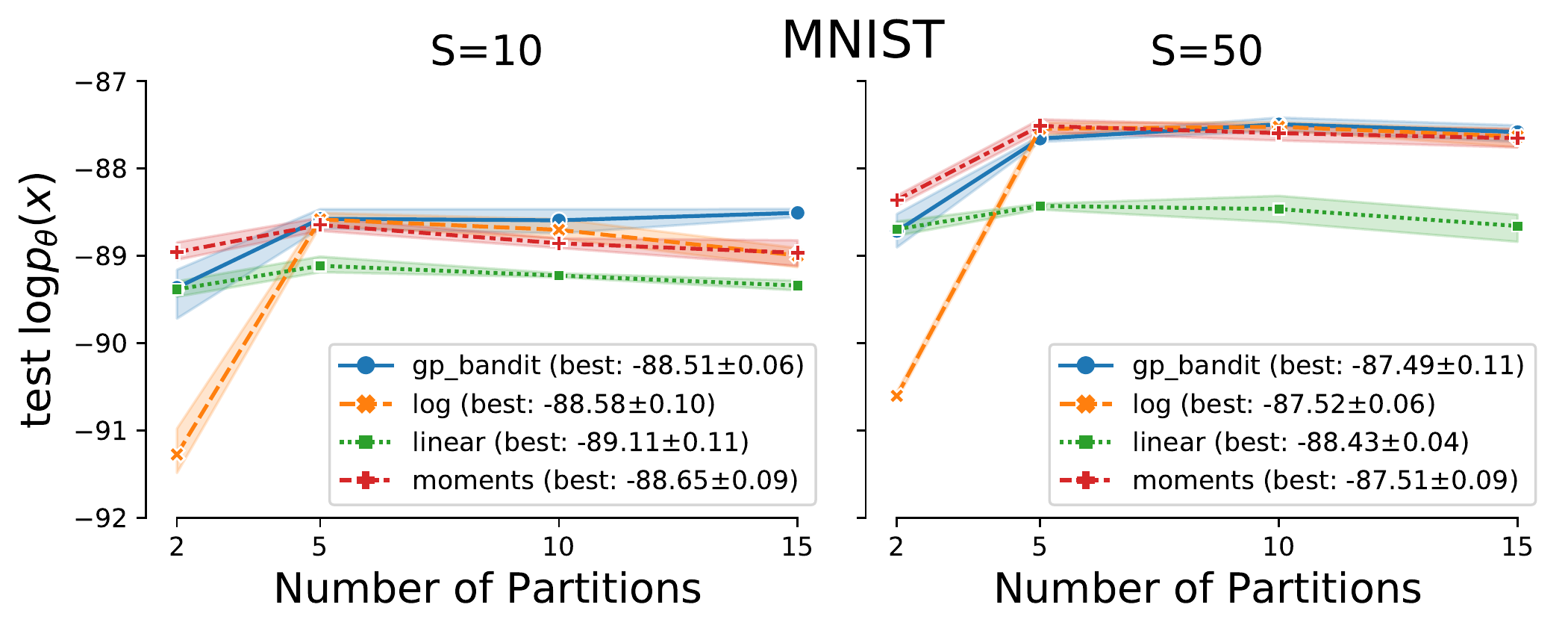}
\includegraphics[width=0.5\columnwidth]{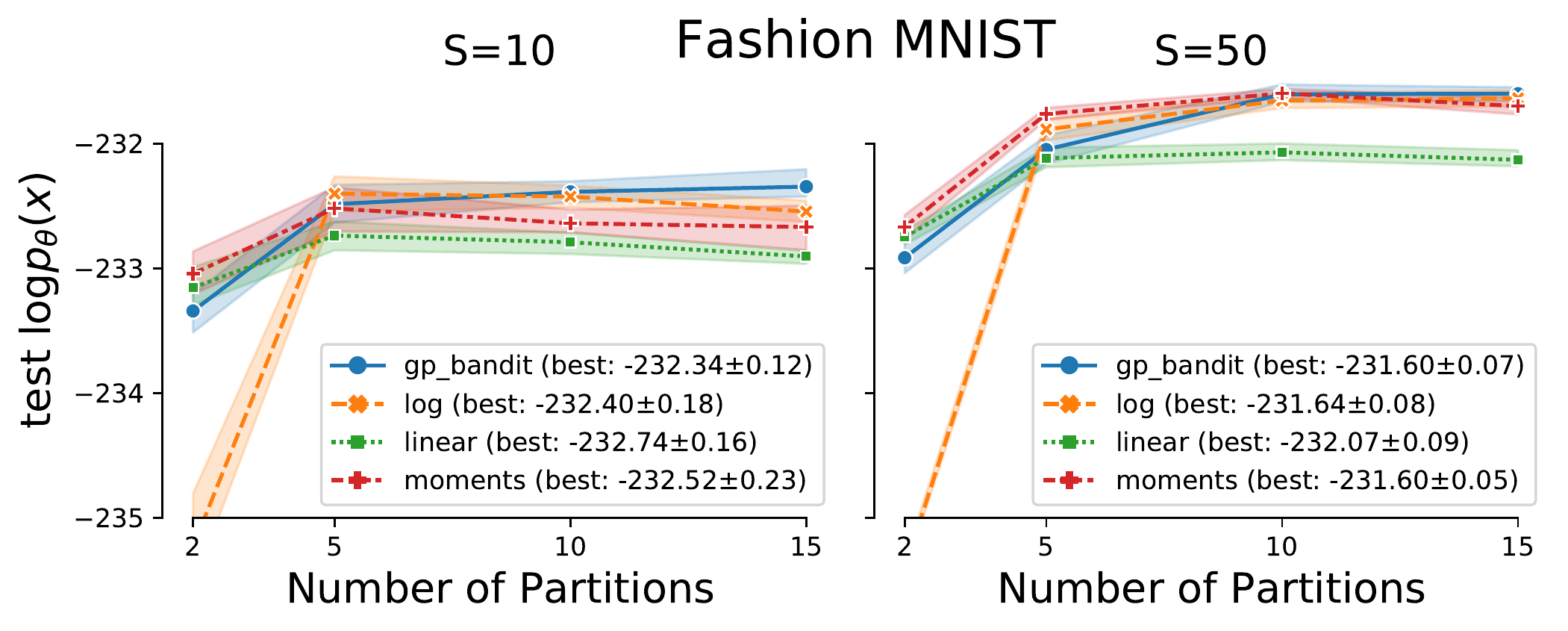}

\includegraphics[width=0.5\columnwidth]{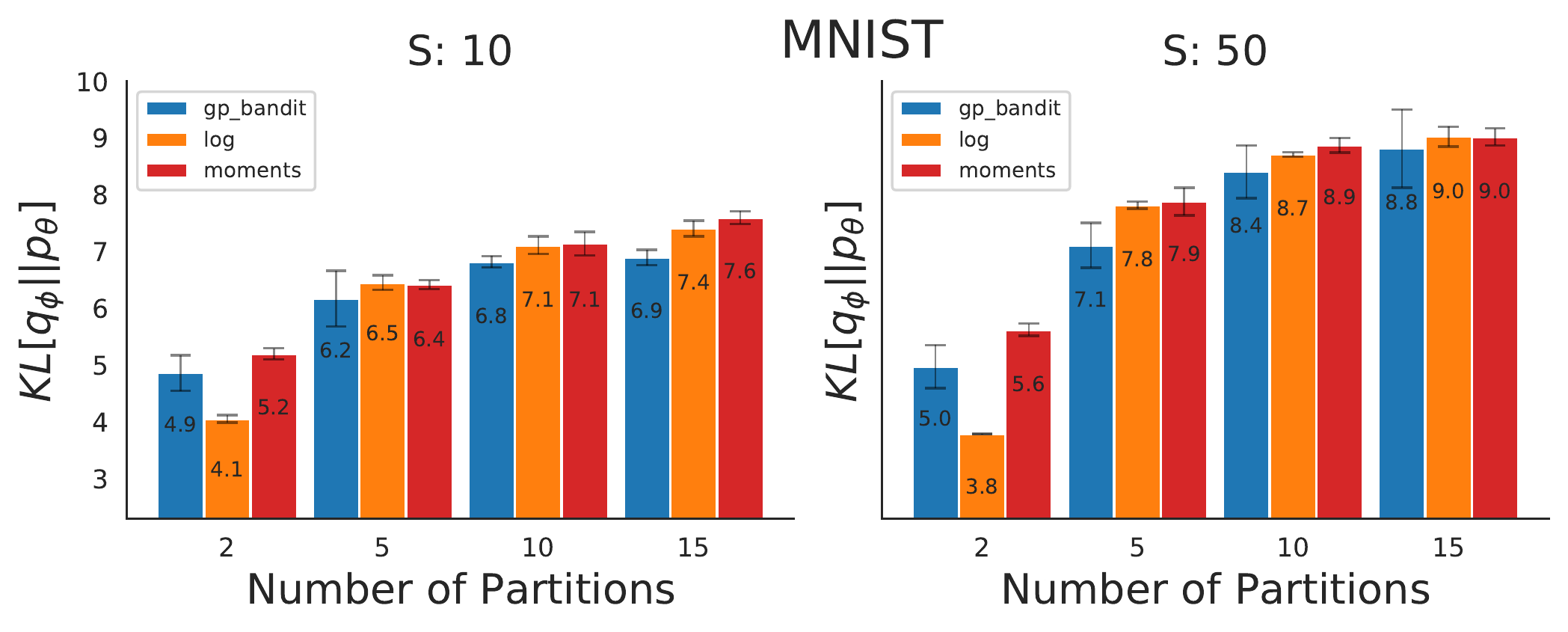}
\includegraphics[width=0.5\columnwidth]{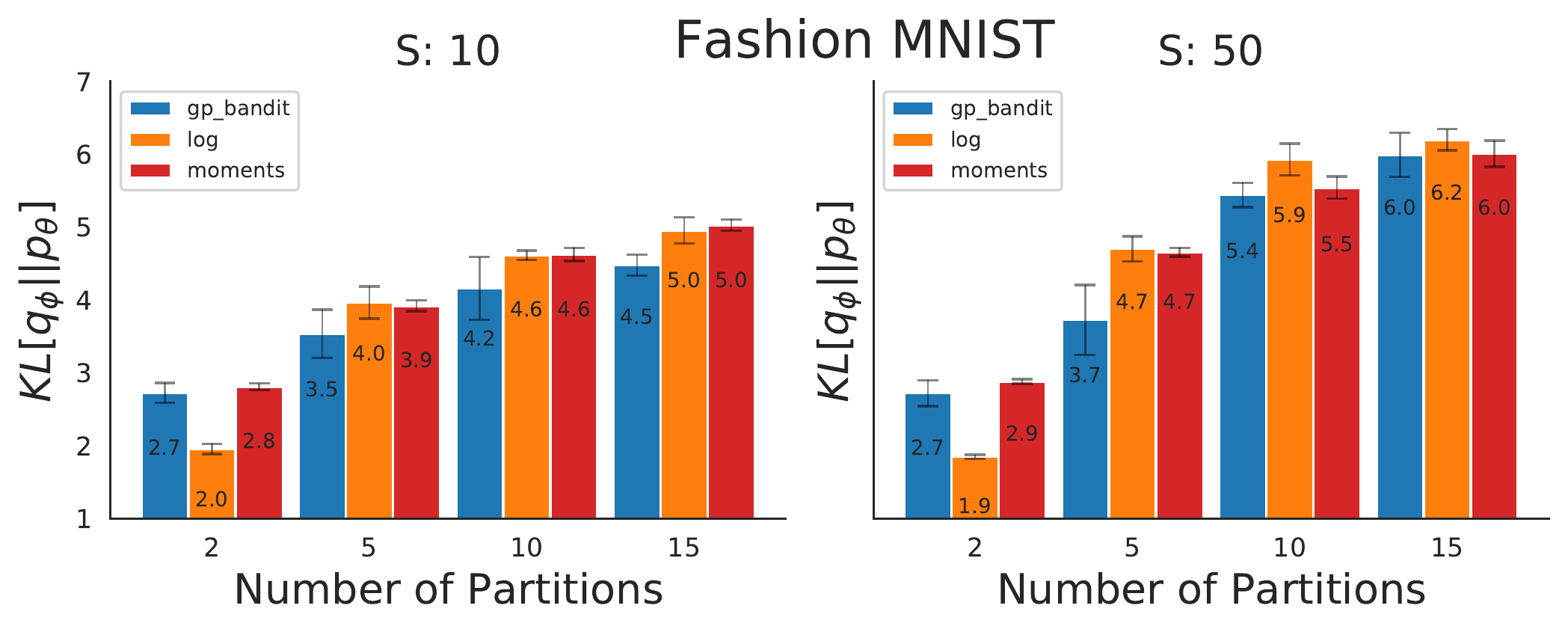}

\caption{Performance comparison for continuous VAE on MNIST and Fashion MNIST. \textit{Top}: we compare model learning performance using test likelihood (higher is better). \textit{Bottom}: 
We compare posterior inference as measured by the test KL divergence (lower is better) against the log and moments baselines.  Although, in general, we find that models with worse $\log p(\vx)$ tend to have lower $D_{KL}$, our GP-bandit schedule provides improvements in both learning and inference. } \label{fig:continuous_VAE}
\end{figure*}
 \subsection{Scheduling Behaviour} \label{sec:Ablation_study} 
We first investigate the behaviour of our time-varying reward function and bandit scheduling.  These experiments highlight the adaptive nature of our algorithm, as we inspect the choice of integration schedule across training epochs for both $d=2$ and $d=5$.  
\paragraph{Time-varying Reward Function:}
In \cref{fig:exmaple_time_reward},  we visualize the mean and variance of our time-varying estimate of the utility function $y_t=f_{t}(\vbeta_{t})+\epsilon_t $ after $2000$, $3000$, and $10,000$ epochs, respectively.  We illustrate the choice of $\vbeta$ for $d=2$, so that $\beta_0 = 0$ is fixed and we can write the reward as $f_t(\beta_1)$.  Colored dots indicate values of $\beta_1$ selected by our bandit algorithm in each round, with the vertical axis reflecting the observed reward $f_t(\beta_1)$ as the change in model evidence $\mathcal{L}$.  

In the first two panels, we observe instances where our bandit prioritizes exploitation, choosing similar, high-reward $\beta_1$ values in neighboring rounds with the same color.  However, note that these $\beta_1$ may not match the highest GP predictive mean for $f_t(\vbeta)$, since the blue line is shown at the final training epoch in a window.  In the final panel, we observe that our time-varying reward function has adapted to have very low variance, since the \gls{TVO} objective changes only slightly near convergence and the choice of $\beta_1$ has little impact.




\paragraph{$\vbeta$ Across Training:} In Figure \ref{fig:beta_overtime}, we visualize bandit choices of $\vbeta$ with $d=5$.   In initial epochs (blue), the GP-bandit algorithm prioritizes exploration before focusing on $\beta_j < 0.5$ in the second phase (orange).  As the \gls{VAE} converges, our algorithm begins to explore $\beta_j$ further from zero (green).

Beyond avoiding the need for an expensive grid search, a primary motivation for our bandit approach is a lack of knowledge about the shape of the integrand.  Using the intuition that $\beta_j$ choices should be concentrated in regions where the integrand is changing quickly in order to obtain accurate Riemann approximations, we can still translate the observed bandit choices of $\vbeta$ into example integrands in the middle (orange) and late (green) stages of training in the bottom panel of Fig. 3.
An integrand that rises steeply away from $\beta = 0$ indicates that $q_{\phi}(\vz|\vx)$ is mismatched to $p_{\theta}(\vz|\vx)$, and the \gls{TVO} might be improved by choosing small $\beta_j$.  As the curve begins to smooth later in training, with a higher proportion of importance samples yielding high likelihood under the generative model, our bandit begins to explore $\beta_j$ closer to $1$.

\subsection{Model Learning and Inference}\label{sec:model_learning_inference}
\paragraph{Continuous VAE: } We present results of training a continuous \gls{VAE} on the MNIST and Fashion MNIST dataset in Figure \ref{fig:continuous_VAE}.  We measure model learning performance using the test log evidence, as estimated by the \textsc{IWAE} bound \cite{burda2015importance} with $5000$ samples per data point.   We also compare inference performance using $D_{\text{KL}}[q_{\phi}(\vz|\vx) \, || p_{\theta}(\vz|\vx)]$, which we calculate by subtracting the test \gls{ELBO} from our estimate of $\log p_{\theta}(\vx)$. 

 \begin{figure*}
\includegraphics[width=0.5\textwidth]{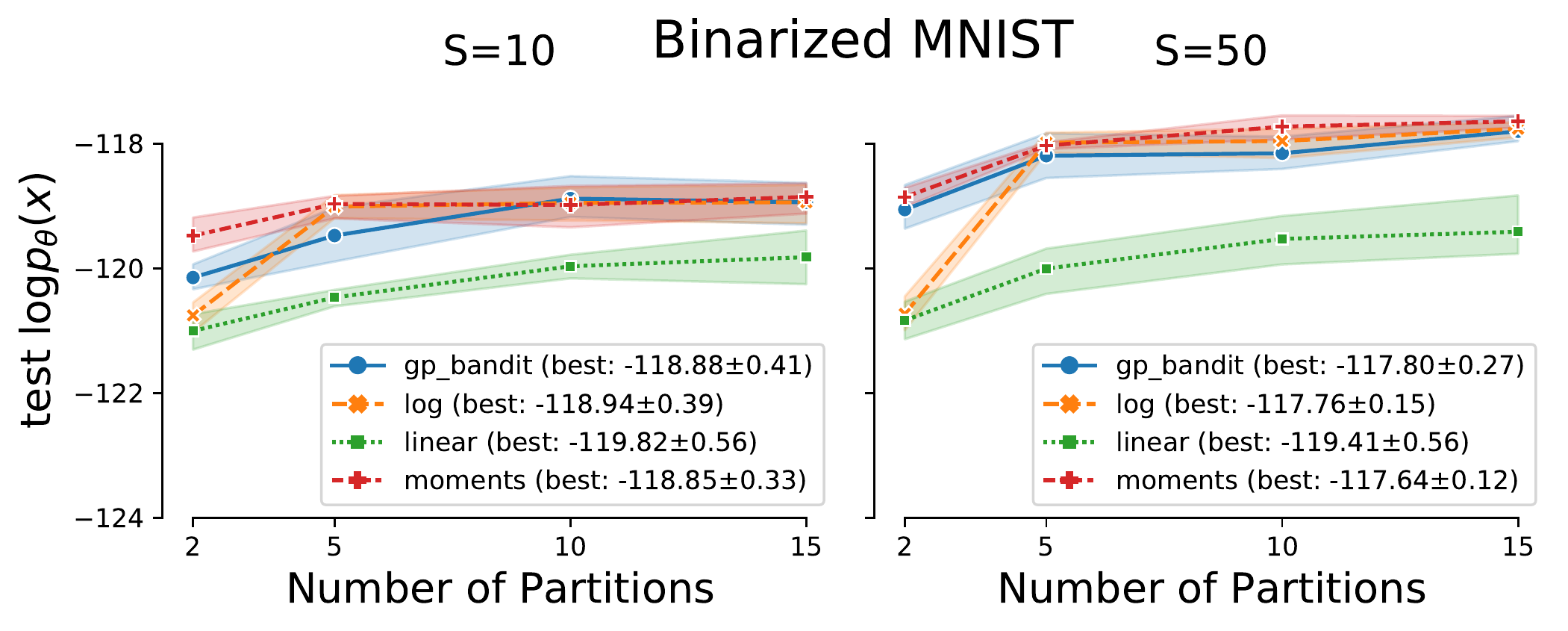}
\includegraphics[width=0.5\textwidth]{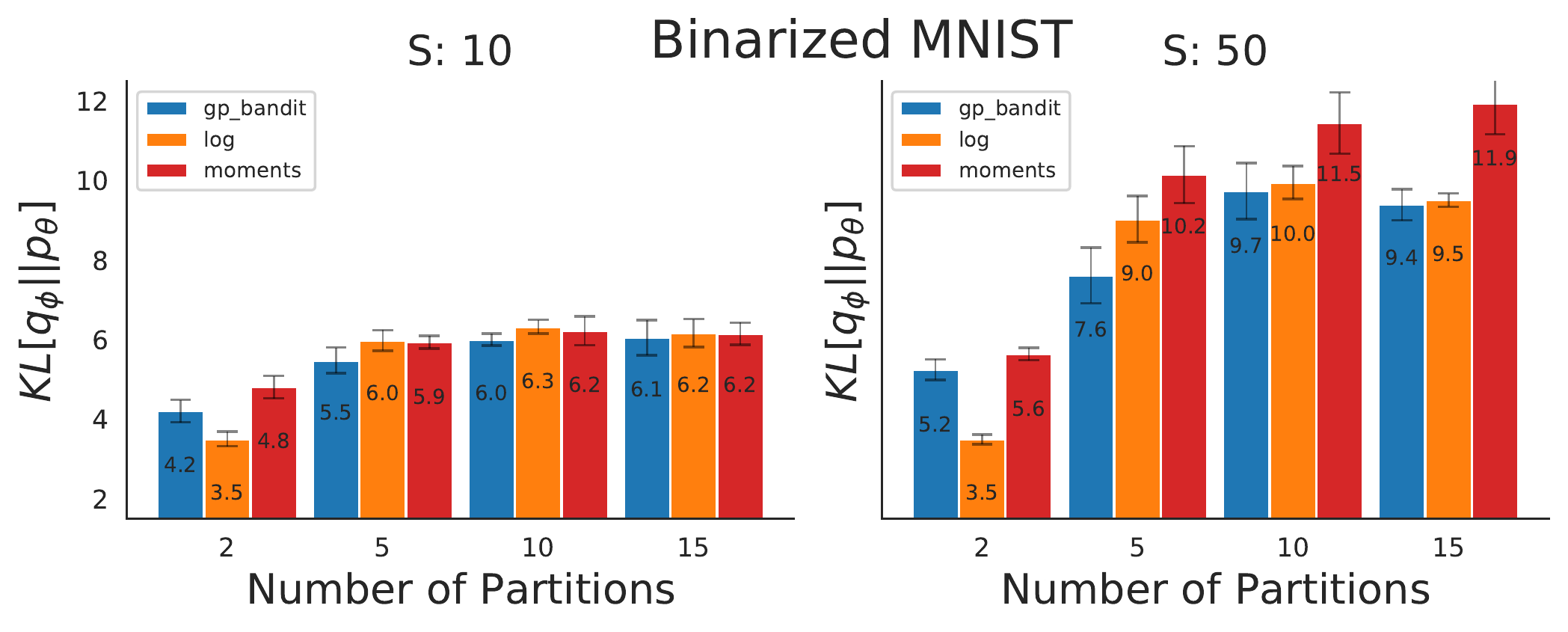}
\caption{Performance comparison in discrete latent variable model using a Sigmoid Belief Network on the binarized MNIST dataset.
Our GP-bandit achieves comparable results to the log and moments schedule in terms model learning (higher $\log p(\bx)$), with better posterior inference (lower $D_{KL}$). \label{fig:discreteVAE_binarizedMNIST}} 
\end{figure*}


For most scenarios in Figure \ref{fig:continuous_VAE}, our GP bandit optimization outperforms baselines with respect to both model learning and inference.  In general, we observe that models with lower model evidence attain lower test KL divergence.  Thus, in comparing inference performance in the bottom panel of Figure \ref{fig:continuous_VAE}, we compare against the log and moment schedules, baselines with comparable test log likelihoods.  It is notable that our approach often achieves better results for both learning (higher $\log p_{\theta}(\vx)$) and inference (lower $D_{\text{KL}}$).  We obtain the highest log evidence with $d=10$ for MNIST and $d=15$ for Fashion MNIST. 

\paragraph{Sigmoid Belief Network: }  We present similar results for learning discrete latent variable models using a Sigmoid Belief Network  \cite{mnih2014neural}.  We show results on binarized MNIST in \cref{fig:discreteVAE_binarizedMNIST}, with binarized Omniglot in \cref{fig:discreteVAE_binarizedOMN} in \cref{sec:discreteVAE}. 
Our GP bandit optimization achieves competitive model learning performance with the log-uniform and moment schedules and better posterior inference across models with comparable $\log p(\vx)$, indicating our GP-bandit schedule can flexibly optimize the \gls{TVO} for various model types.

\section{Conclusion \label{sec:Conclusion-and-Future}}
We have presented a new approach for automated selection of the integration schedule for the Thermodynamic Variational Objective.  Our bandit framework optimizes a reward function that is directly linked to improvements in the generative model evidence over the course of training the model parameters.  We show theoretically that this procedure asymptotically minimizes the regret as a function of the choice of schedule.  Finally, we demonstrated that the proposed approach empirically outperforms existing schedules in both model learning and inference for discrete and continuous generative models.




Our GP bandit optimization offers a general solution to choosing the integration schedule in the \gls{TVO}. However, our algorithm, as well as all other existing schedules, still rely on the number of partitions $d$ as a hyperparameter which is fixed over the course of the training.   Incorporating the adaptive selection of $d$ into our bandit optimization remains an interesting direction for future work.

\section{Broader Impact}
Our research can be widely applied for variational inference in deep generative models, including variational autoencoders with autoregressive decoders and normalizing flows.  Variational inference, and Bayesian methods more generally, have broad applications spanning science and engineering, from epidemiology~\cite{wood2020planning} to particle physics~\cite{baydin2019efficient}. Our methodological contributions for variational inference may find broader impact through improved modelling in these disparate domains.  However, our method is general in nature, so domain-specific applications should further consider implications for deployment in the real-world.

\section{Acknowledgements}
 
VM acknowledges the support of the Natural Sciences and Engineering Research Council of Canada (NSERC) under award number PGSD3-535575-2019 and the British Columbia Graduate Scholarship, award number 6768. 
VM/FW acknowledge the support of the Natural Sciences and Engineering Research Council of Canada (NSERC), the Canada CIFAR AI Chairs Program, and the Intel Parallel Computing Centers program. 
RB acknowledges support from the Defense Advanced Research Projects Agency (DARPA) under award FA8750-17-C-0106.

This material is based upon work supported by the United States Air Force Research Laboratory (AFRL) under the Defense Advanced Research Projects Agency (DARPA) Data Driven Discovery Models (D3M) program (Contract No. FA8750-19-2-0222) and Learning with Less Labels (LwLL) program (Contract No.FA8750\-19\-C\-0515). 
Additional support was provided by UBC's Composites Research Network (CRN), Data Science Institute (DSI) and Support for Teams to Advance Interdisciplinary Research (STAIR) Grants. 
This research was enabled in part by technical support and computational resources provided by WestGrid (\url{https://www.westgrid.ca/}) and Compute Canada (\url{www.computecanada.ca}).

\bibliography{vunguyen}


\clearpage


\appendix
\clearpage 

\section{Experimental Setup} \label{app:ExperimentalSetup}
\paragraph{Dataset Description}
The discrete and continuous VAE literature use slightly different training procedures. For continuous VAEs, we follow the sampling procedure described in footnote 2, page 6 of \citet{burda2015importance}, and sample binary-valued pixels with expectation equal to the original gray scale $28 \times 28$ image. We split MNIST~\cite{lecun1998gradient} and Fashion MNIST~\cite{xiao2017fashion} into 60k training examples and 10k testing examples across $10$ classes.

For Sigmoid Belief Networks, we follow the procedure described by \citet{mnih2016variational} and use 50k training examples and 10k testing examples (the remaining 10k validation examples are not used) with the binarized MNIST~\cite{salakhutdinov2008quantitative} dataset.

\paragraph{Training Procedure}
All models are written in PyTorch and trained on GPUs. For each scheduler, we train for $10,000$ epochs using the Adam optimizer \cite{kingma2014adam} with a learning rate of $10^{-3}$, and minibatch size of $1000$. All weights are initialized with PyTorch's default initializer. For the neural network architecture, we use two hidden layers of $[100,25]$ nodes.

\paragraph{Reward Evaluation }
To obtain the bandit feedback in Eq. \eqref{eq:reward_function}, we use a fixed, linear schedule with $d=50$ for calculating $\mathcal{L}_t$ with Eq. \eqref{eq:left_Riemann_sum}.  This yields a tighter $\log p_{\theta}(\vx)$ bound, decouples reward function evaluation from model training and schedule selection in each round, and is still efficient using \gls{SNIS} in Eq. \eqref{eq:snis}.  We limit the value of $d$ for \gls{TVO} training following observations of deteriorating performance in \cite{masrani2019thermodynamic}.

\section{GP kernels and treatment of GP hyperparameters}\label{app:GP}

We present the GP kernels and treatment of GP hyperparameters for
the black-box function $f$.

We use the exponentiated quadratic (or squared exponential) covariance function for input hyperparameter $k_{\beta}(\vbeta,\vbeta')=\exp\left(-\frac{||\vbeta-\vbeta'||^{2}}{2\sigma_{\beta}^{2}}\right)$
and a time kernel $k_{T}(t,t')=(1-\omega)^{\frac{|t-t'|}{2}}$ where
the observation $\bbeta$ and $t$ are normalized to $[0,1]^{d}$ and
the outcome $y$ is standardized $y\sim\mathcal{N}\left(0,1\right)$
for robustness. As a result, our product kernel becomes
\begin{align*}
k\left([\bbeta,t],[\bbeta',t']\right) & =k(\bbeta,\bbeta')\times k(t,t')=\exp\left(-\frac{||\bbeta-\bbeta'||^{2}}{2\sigma_{\beta}^{2}}\right)(1-\omega)^{\frac{|t-t'|}{2}}.
\end{align*}

The length-scales $\sigma_{\beta}$ is estimated from the data indicating the variability of the function with regards to the hyperparameter input $\bx$ and number of training iterations $t$. Estimating appropriate values for them is critical as this represents the GP's prior regarding the sensitivity of performance w.r.t. changes in the number of training iterations and hyperparameters. We note that previous works have also utilized the above product of spatial and temporal covariance functions for different settings \cite{Krause_2011Contextual, bogunovic2016time,boil}.


We fit the GP hyperparameters by maximizing their posterior probability (MAP), $p\left(\sigma_{l},\omega\mid \bbeta,\mathbf{t},\by\right)\propto p\left(\sigma_{l},\omega,\bbeta,\mathbf{t},\by\right)$, which, thanks to the Gaussian likelihood, is available in closed form as \cite{Rasmussen_2006gaussian}
\begin{align}
\ln p\left(\by,\bbeta,\mathbf{t},\sigma_l,\omega\right)= & \frac{1}{2}\by^{T}\left(\bK+\sigma^{2}\idenmat_{N}\right)^{-1}\by-\frac{1}{2}\ln\left|\bK+\sigma^{2}\idenmat_{N}\right|+\ln p_{\textrm{hyp}}\left(\sigma_{x},\omega\right)+\textrm{const}\label{eq:MarginalLLK}
\end{align}
 where $\idenmat_{N}$ is the identity matrix in dimension $N$ (the number of points in the training set), and $p_{\textrm{hyp}}(\sigma_{l},\omega)$ is the prior over hyperparameters, described in the following.

We maximize the marginal likelihood in Eq. (\ref{eq:MarginalLLK}) to select the suitable lengthscale parameter $\sigma_l$, remembering-forgetting trade-off $\omega$, and noise variance $\sigma^2_f$.

Optimizing Eq. (\ref{eq:MarginalLLK}) involves taking the derivative w.r.t. each variable, such as $\frac{\partial \ln p\left(\by,\bbeta,\mathbf{t},\sigma_l,\omega\right)}{\partial \omega}=\frac{\partial \ln p\left(\by,\bbeta,\mathbf{t},\sigma_l,\omega\right)}{\partial \bK} \times \frac{\partial \bK}{\partial k(t,t')} \times \frac{\partial k(t,t')}{\partial \omega}$.
 While the derivatives of $\sigma_l$ and $\sigma^2_f$ are standard and can be found in \cite{Rasmussen_2006gaussian}, we present the derivative w.r.t. $\omega$  as follows
\begin{align}
 \frac{\partial k(t,t')}{\partial{\omega}} = -v \left(1-\omega \right)^{v-1} \text{where } v=|t-t'|/2.
\end{align}

We optimize Eq. (\ref{eq:MarginalLLK}) with a gradient-based optimizer, providing the analytical gradient to the algorithm. We start the optimization from the previous hyperparameter values $\theta_{prev}$.  If the optimization fails due to numerical issues, we keep the previous value of the hyperparameters.

\section{Proof of Theorem \ref{thm:theorem_sublinear_regret}} \label{app:theory}
Our use of the TVGP within the TVO setting requires no problem specific modifications compared to the general formulation in Bogunovic.  As such, the proof of Theorem 1 closely follows the proof of Theorem 4.3 in \citet{bogunovic2016time} App. C. with time kernel $k_{\textrm{T}}\left(i,j\right)=\left(1-\omega\right)^{\frac{|i-j|}{2}}$.
At a high level, their proof proceeds by partitioning the $T$ random functions into blocks of length $\tilde{N}$, and bounding each using Mirsky's theorem. Referring to ~\cref{tab:theory_notation} for notation, this results in a bound on the maximum mutual information
\begin{align}
    \tilde{\gamma}_{\tilde{N}} \le \left(\frac{T}{\tilde{N}} + 1\right)\left(\gamma_{\tilde{N}} + \tilde{N}^{3}\omega\right), \label{eq:max_mut}
\end{align}
which leads directly to their bound on the cumulative regret (cf. App C.2 in \cite{bogunovic2016time}). Our contribution is to recognize we can achieve a tighter bound on the maximum mutual information with an application Cauchy Schwarz and Jensen's inequality.
\begin{proof}
Beginning from \citet{bogunovic2016time} Eq. (58), we have
\begin{align}
    \tilde{\gamma}_{\tilde{N}} & \le \gamma_{\tilde{N}} + \sum_{i=1}^{\tilde{N}}\log\left(1 + \Delta_i\right) & & \\
    \tilde{\gamma}_{\tilde{N}} & \le \gamma_{\tilde{N}} + \tilde{N} \log\left(1 + \frac{1}{\tilde{N}}\sum_{i=1}^{\tilde{N}} \Delta_i\right) & &\text{Jensen's inequality} \\
    \tilde{\gamma}_{\tilde{N}} & \le \gamma_{\tilde{N}} + \tilde{N} \log\left(1 + \frac{1}{\sqrt{\tilde{N}}}\sqrt{\sum_{i=1}^{\tilde{N}} \Delta_i^2}\right) & &\text{Cauchy-Schwartz} \\
    \tilde{\gamma}_{\tilde{N}} & \le \gamma_{\tilde{N}} + \tilde{N} \log\left(1 + \tilde{N}^{3/2}\omega\right) & &\sum_{i=1}^{\tilde{N}} \Delta_i^2 \le \tilde{N}^4\omega^2\\
    \tilde{\gamma}_{\tilde{N}} & \le \gamma_{\tilde{N}} + \tilde{N}^{5/2} \omega & &\log(1 + x) \le x\label{eq:result}
\end{align}
This bound is tighter than \cite{bogunovic2016time} Eq. (60) $(\tilde{N}^{5/2}\le\tilde{N}^{3})$, where the latter was achieved via a simple constrained optimization argument. Using \cref{eq:result}, Theorem 1 follows using identical arguments as in \cite{bogunovic2016time}.
\end{proof}

\begin{table*}
    \caption{Supporting notations in regret analysis\label{tab:Notation-List-regret}. We use notation to similar to Appendix C of \citet{bogunovic2016time} when possible.}\label{tab:theory_notation}
    \centering{}%
    \begin{tabular}{llll}
    \toprule
    Parameter & Domain & Meaning & \tabularnewline
    \midrule
    $\omega$ & scalar, $(0, 1)$ & Remembering-forgetting trade-off parameter ($\epsilon$ in \cite{bogunovic2016time}) & \\ \addlinespace[0.1cm]
    ${\bf f}_{T}$ & vector, $\mathbb{R}^T$ & Vector of $T$ function evaluations from $f$, ${\bf f_T} := [f(x_1),...,f(x_T)]^T$. & \\ \addlinespace[0.1cm]
    ${\bf \tilde{f}}_{T}$ & vector, $\mathbb{R}^T$ & (\textit{time-varying case}) Vector of $T$ function evaluations from $f_{1:T}$, \\ && ${\bf \tilde{f}}_T := [f_1(x_1),...,f_T(x_T)]^T$& \\ \addlinespace[0.1cm]
    $I(\by_{T}; {\bf f}_{T})$ & scalar, $\mathbb{R}^+$ &  The mutual information in ${\bf f}_{T}$  after revealing $\by_{T} = {\bf f}_{T} + {\bf \epsilon}$.\\ && For a GP with covariance function  $\bm{K}_T$, $I(\by_{T}; {\bf f}_{T}) = \frac{1}{2}\log|\idenmat + \sigma^{-2}\bm{K}_T|$& \\ \addlinespace[0.1cm]
    $\tilde{I}(\by_{T}; {\bf \tilde{f}}_{T})$ & scalar, $\mathbb{R}^+$ & (\textit{time-varying case}) Mutual information $\tilde{I}(\by_{T}; {\bf \tilde{f}}_{T}) = \frac{1}{2}\log|\idenmat + \sigma^{-2}\tilde{\bm{K}}_T|$,\\ && where $\tilde{\bm{K}}_T$ is a covariance function that incorporates time kernel $k_{\textrm{T}}(i, j)$ & \\ \addlinespace[0.1cm]
    $\gamma_{T}$ & scalar, $\mathbb{R}^+$ &  The maximum information gain $\gamma_{T} := \max\limits_{x_1, ..., x_T}I(\by_{T}; {\bf f}_{T})$ after $T$ rounds& \\ \addlinespace[0.1cm]
    $\tilde{\gamma}_{T}$ & scalar, $\mathbb{R}^+$ &  (\textit{time-varying case}) The analogous time-varying maximum information \\ && gain $\tilde{\gamma}_{T} := \max\limits_{x_1, ..., x_T}\tilde{I}(\by_{T}; {\bf \tilde{f}}_{T})$& \\\addlinespace[0.1cm]
    $\tilde{N}$ & scalar, $\mathbb{R}^+$ & An artifact of the proof technique used by \cite{bogunovic2016time}. The $T$ time steps \\ && are partitioned into blocks of length $\tilde{N}$& \\\addlinespace[0.1cm]
     &  &  & \tabularnewline
    \end{tabular}
\end{table*}

\section{Additional Experiments and Ablation Studies} \label{app:experiments}
We present additional experiments on a Probabilistic Context Free Grammar (PCFG) model and Sigmoid Belief Networks in \S \ref{app:pcfg} and \S \ref{sec:discreteVAE}, a wall-clock time benchmark in \S \ref{app:wallclock}, ablation studies in \S \ref{app:ablation_study}, and additional training curves in \S \ref{app:training_curves}.

\subsection{Training Probabilistic Context Free Grammar} \label{app:pcfg}

\begin{figure*}[t]
\includegraphics[width=1\columnwidth]{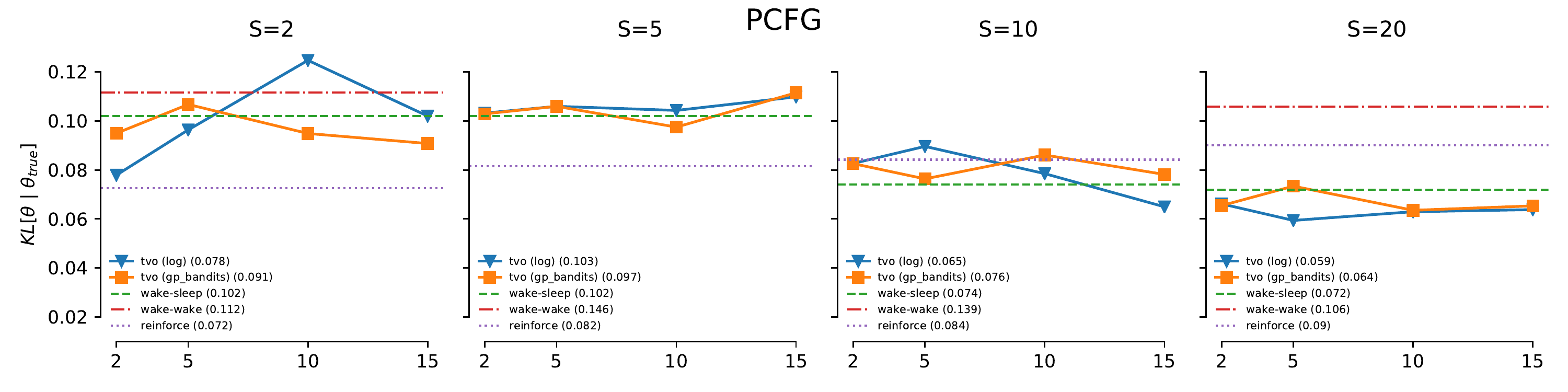}
\includegraphics[width=1\columnwidth]{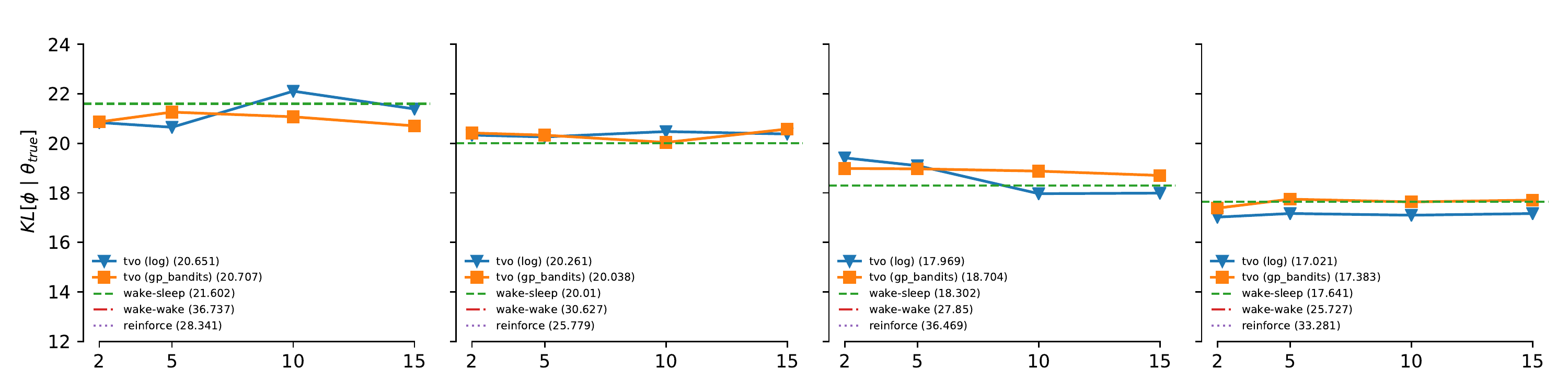}
\includegraphics[width=1\columnwidth]{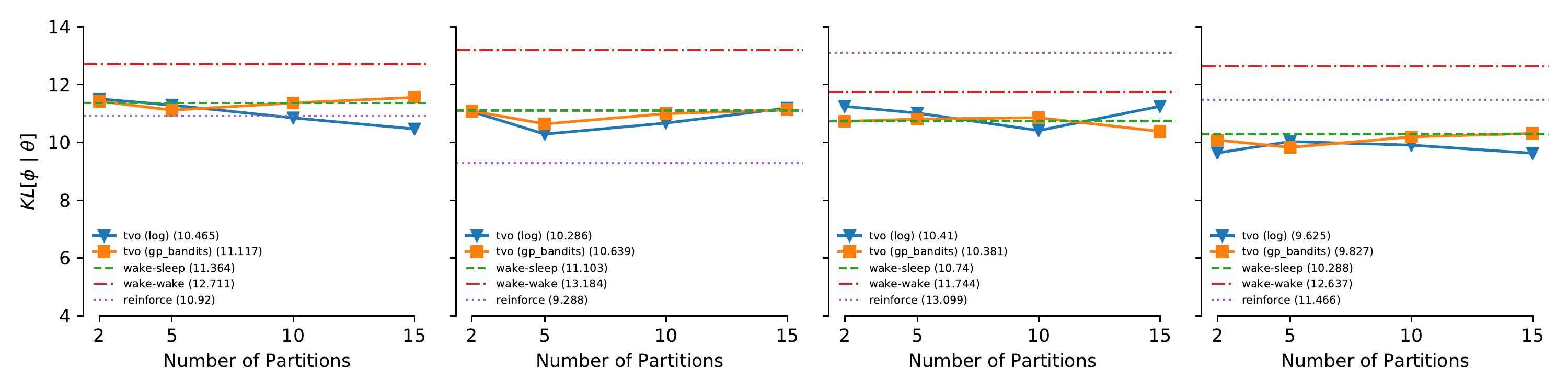}
\caption{Evaluation of model learning in a PCFG, where $\theta$ is the set of probabilities associated with each production rule in the grammar, and $\phi$ is an RNN which generates the conditional probability of a parse tree given a sentence. GP-bandits (ours) is comparable to the baseline log schedule and less sensitive to number of partitions, as evaluated by the KL divergence between learned and true model parameters (top row). Inference network learning is evaluated by the KL divergence between $q_{\phi}(\vz|\vx)$ and $p_{\theta_{\text{true}}}(\vz|\vx)$ (middle row) and $p_{\theta}(\vz|\vx)$ (bottom row). We compare against \textsc{reinforce}, \textsc{wake-wake}, and \textsc{wake-sleep}, where some baselines aren't shown due to being out of range. At $S=20$, TVO with log and GP-bandits schedule outperforms \textsc{reinforce}, \textsc{wake-wake}, and \textsc{wake-sleep} both in terms of the quality of the generative model (top row, right) and inference network (middle row, right) for all $S \in \{2, 5, 10, 20\}$. } \label{fig:pcfg}
\vspace{-5pt}
\end{figure*}

In order to evaluate our method outside of the Variational Autoencoder framework, we consider model learning and amortized inference in the probabilistic context-free grammar setting described in Section 4.1 of \citet{le2019revisiting}. Here $p_{\theta}(\vx, \vz) = p(\vx|\vz)p_\theta(\vz)$, where $p_\theta(\vz)$ is a prior over parse trees $\vz$, $p(\vx|\vz)$ is a soft relaxation of the $\{0, 1\}$ likelihood which indicates if sentence $\vx$ matches the set of terminals (i.e the sentence) produced by $\vz$, and $\theta$ is the set of probabilities associated with each production rule in the grammar. The inference network $\phi$ is a recurrent neural network which outputs $q_\phi(\vz|\vx)$, the conditional distribution of a parse tree given an input sentence. We use the \textit{Astronomers} PCFG considered by \citet{le2019revisiting}, and therefore have access to the ground truth production probabilities $\theta_{\text{true}}$, which we will use to evaluate the quality of our learned model $\theta$.

We compare the TVO with GP-bandit and log schedules against \textsc{reinforce}, \textsc{wake-wake}, and \textsc{wake-sleep}, where \textsc{wake-wake} and \textsc{wake-sleep} use data from the true model ${\vx \sim p_{\theta_{\text{true}}}(\vx)}$ and learned model ${\vx \sim p_{\theta}(\vx)}$ respectively.
For each run, we use a batch size of $2$ and train for $2000$ epochs with Adam using default parameters.
For all KL divergences (see caption in Figure~\ref{fig:pcfg}), we compute the median over $20$ seeds and then plot the average over the last $100$ epochs.

As observed by \citet{le2019revisiting}, sleep-$\phi$ updates avoid the deleterious effects of the SNIS bias and is therefore preferable to wake-$\phi$ updates in this context. Therefore for all runs we use the \gls{TVO} to update $\theta$, and use sleep-$\phi$ to update $\phi$. Sleep-$\phi$ is a special case of the \gls{TVO} (cf.~\citet{masrani2019thermodynamic} Appendix G.1). 

In Figure~\ref{fig:pcfg} we see that both GP-bandits and log schedules have comparable performance in this setting, with \gls{TVO}-log, $S=20$ achieving the lowest $\text{KL}[p_{\theta} || p_{\theta_{\text{true}}}]$ and $\text{KL}[q_{\phi}(\vz|\vx) || p_{\theta_{\text{true}}}(\vz|\vx) ]$ across all trials. $\text{KL}[q_{\phi} || p_{\theta_{\text{true}}}]$ is a preferable metric to $\text{KL}[q_{\phi} || p_{\theta}]$ because the former does not depend on the quality of the learned model. We also note that GP-bandits appears to be less sensitive to the number of partitions than the log schedule.

\subsection{Training Sigmoid Belief Network on Binarized Omniglot} \label{sec:discreteVAE}
\begin{figure*}
\includegraphics[width=0.49\columnwidth]{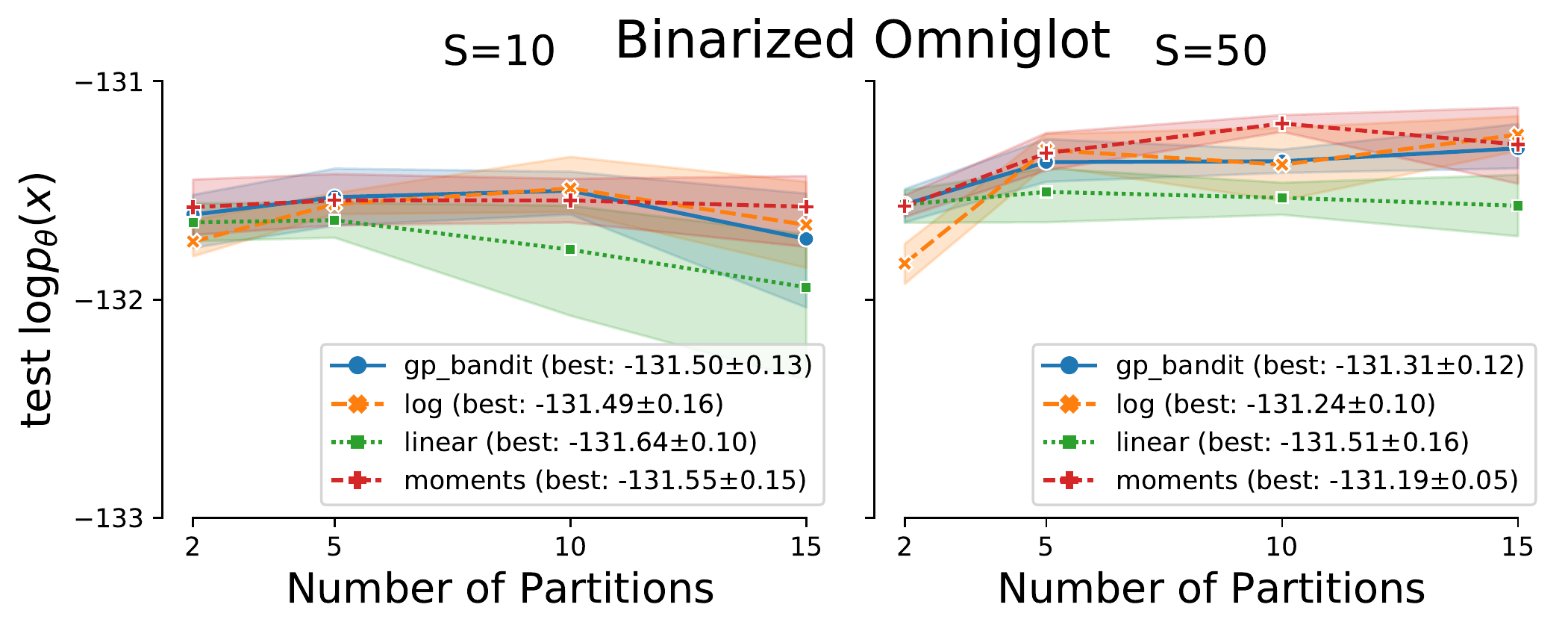}
\includegraphics[width=0.49\columnwidth]{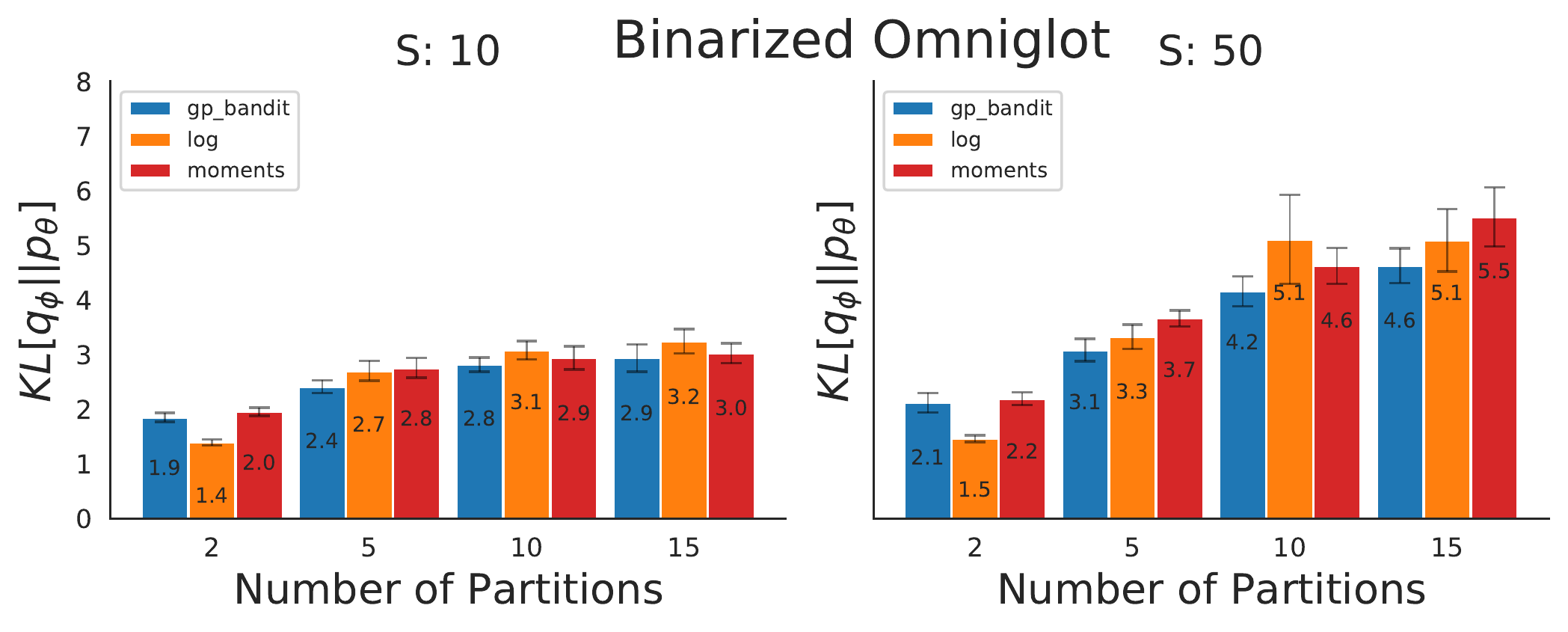}

\caption{Performance of the Sigmoid Belief Network described in \S \ref{sec:model_learning_inference} on the binarized omniglot dataset. The GP-bandit schedule at $d=15, S=50$ outperforms all baselines in terms of model and inference network learning. \label{fig:discreteVAE_binarizedOMN}}
\end{figure*}

We train the Sigmoid Belief Network described in \S \ref{sec:model_learning_inference} on the binarized Omniglot dataset. Omniglot \cite{lake2013one} has $1623$ handwritten characters across $50$ alphabets. We manually binarize the omniglot\cite{lake2015human} dataset by sampling once according to procedure described in \cite{burda2015importance}, and split the binarized omniglot dataset into $23,000$ training and $8,070$ test examples.
Results are shown in Figure \ref{fig:discreteVAE_binarizedOMN}. At $S=50$, GP-bandit achieves similar model learning but better inference compared to log scheduling.

\subsection{Wall-clock time Comparison} \label{app:wallclock}
We benchmark the wall-clock time of our GP-bandit schedule against the cumulative wall-clock time of the grid-search log schedule. For both schedules we train a VAE on the Omniglot dataset for $S=10$ and $5000$ epochs. For the log schedule, we run the sweep ran by \citet{masrani2019thermodynamic} (cf. section 7.2), i.e. $20$ $\beta_1$ linearly spaced between $[10^{-2}, 0.9]$ for $d={2,...,6}$, for a total of $100$ runs. For a fair comparison against the log schedule, we loop over $d={2,...,6}$ for our GP bandits because $d$ is unlearned, for a total of five runs. We note that each run of the GP bandits schedule includes learning the GP hyperparameters as described in \cref{app:GP}.
For both schedules, we take the best $\log p(x)$ and corresponding KL divergence, and plot the cumulative run time across all runs. The results in Table \ref{table:wallclock} show that the GP-bandits schedule does comparable to the grid searched log schedule (log likelihood:  $-110.72$ vs $-110.99$) while requiring significantly less cumulative wall-clock time ($10$ hrs vs $178$ hrs).

\subsection{Ablation studies} \label{app:ablation_study}
\paragraph{Ablation study between GP-bandit and random search.} 
To demonstrate that our model can leverage useful information from past data, we compare against the Random Search picks the integration schedule uniformly at random.

\begin{figure*}
\includegraphics[width=0.5\columnwidth]{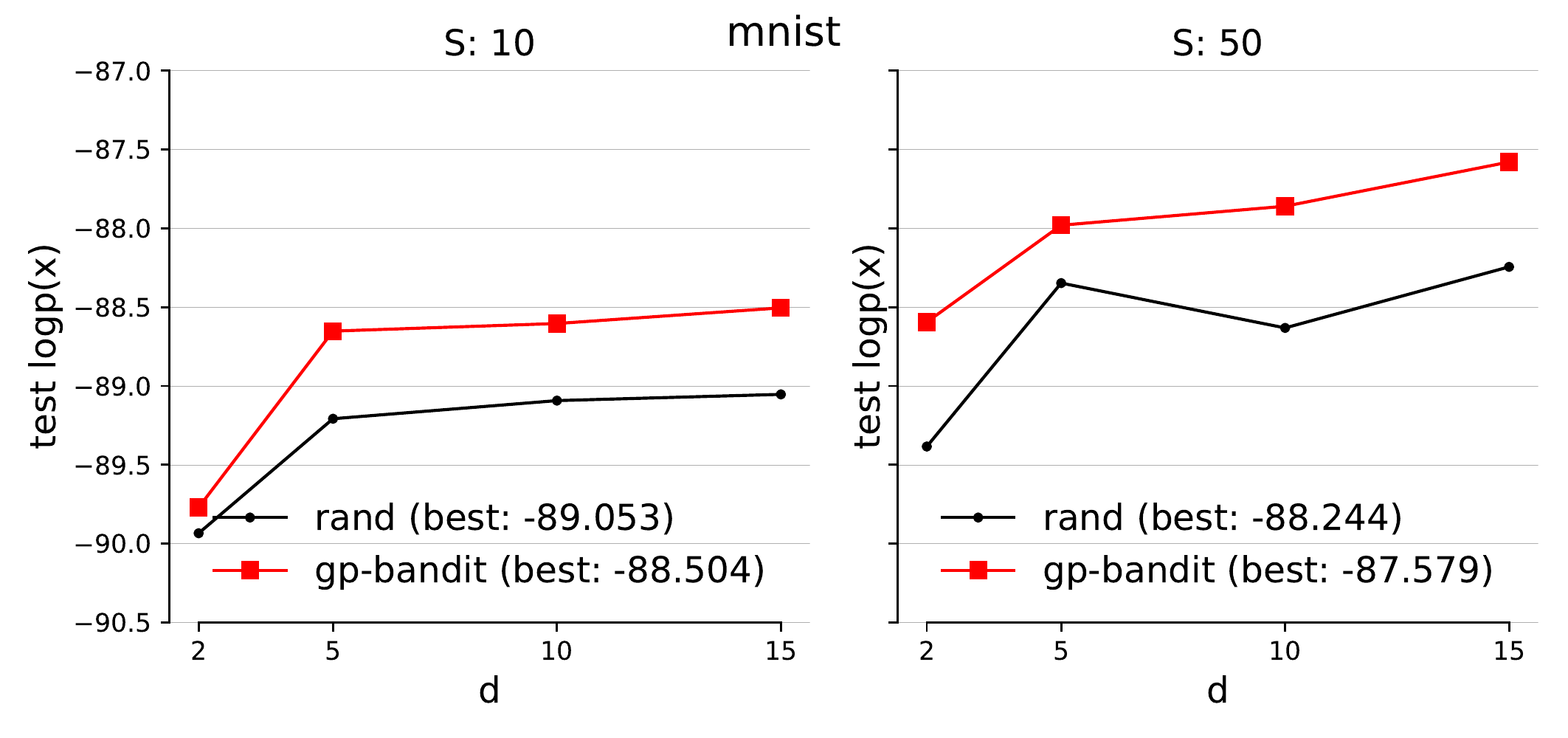}
\includegraphics[width=0.5\columnwidth]{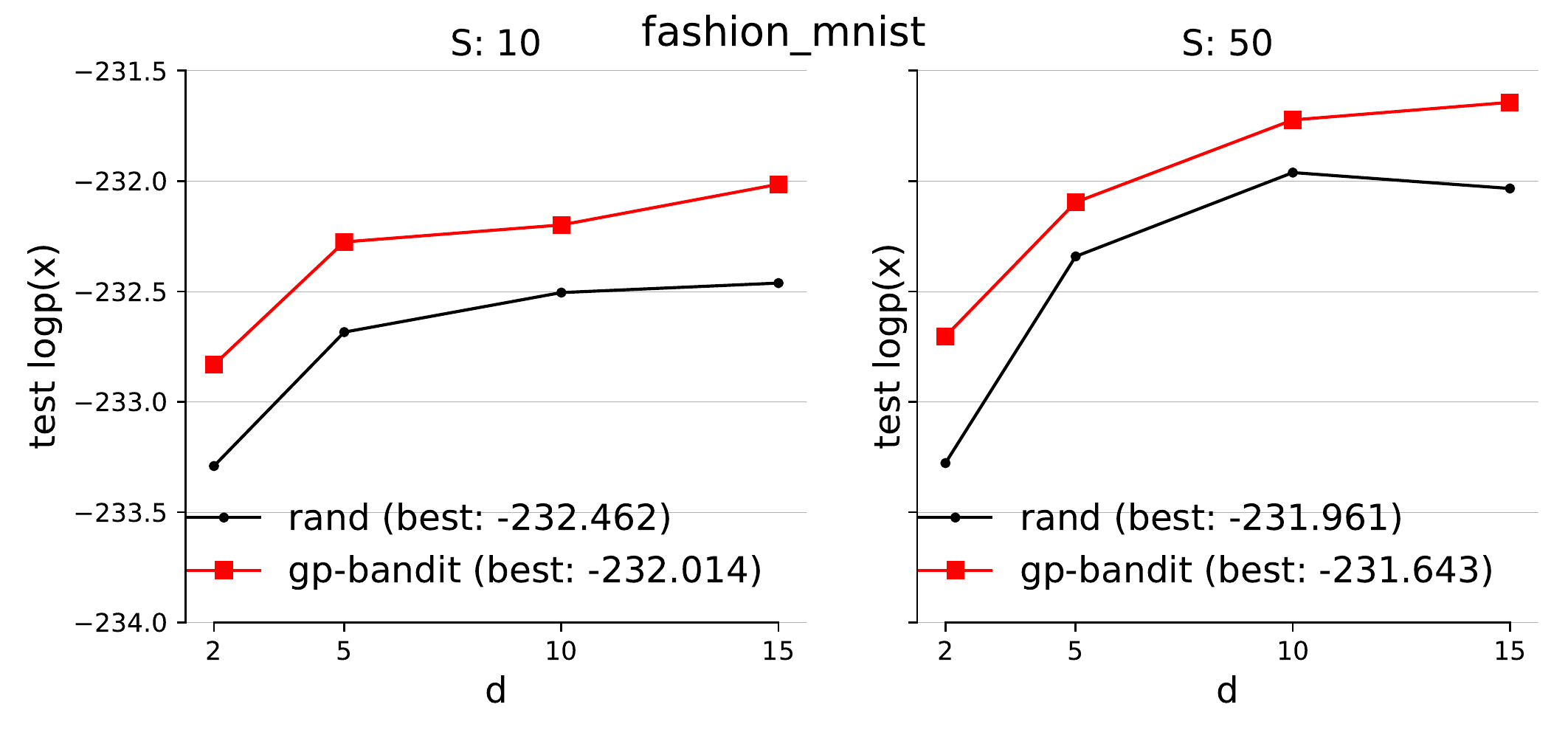}

\caption{We compare the performance between our GP-bandit against the Random search (Rand) baseline which uniformly generates the integration schedules $\vbeta_t$. The GP-bandit schedule outperforms the random counterpart by using information obtained from previous choices, as described in \cref{alg:TVO_GPTV_highlevel}.
\label{fig:ablation_random}}
\end{figure*}

We present the results in Figure \ref{fig:ablation_random} using MNIST (left) and Fashion MNIST (right). We observe that our GP-bandit clearly outperforms the Random baseline. The gap is generally increasing with larger dimension $d$, e.g., $d=15$ as  the search space grows exponentially with the dimension.

\paragraph{Ablation study between permutation invariant GPs.} 

We compare our GP-bandit model using two versions of (1) non-permutation invariant GP and (2) permutation invariant GP in Table \ref{table:ablation}.

Our permutation invariant GP does not need to add all permuted observations into the model, but is still capable of generalizing. The result in Table \ref{table:ablation} confirms that if we have more samples to learn the GP, such as using larger epochs budget $T$, the two versions will result in the same performance. On the other hand, if we have limited number of training budgets, e.g., using lower number of epochs, the permutation invariant GP will be more favorable and outperforms the non-permutation invariant. In addition, the result suggests that for higher dimension $d=15$ (number of partitions) our permutation invariant GP performs consistently better than the counterpart.

\begin{table}
\caption{Wallclock time of the GP-bandit schedule compared to the grid-search of \cite{masrani2019thermodynamic} for the log schedule. GP-bandit approach achieves a competitive test log likelihood and lower KL divergence compared with the grid-searched log schedule, but requires significantly lower cumulative run-time. } \label{table:wallclock}
\begin{centering}
\begin{tabular}{ccccc}
\toprule
& best $\log p(x)$ & best kl & number of runs & cumulative run time (hrs) \\ \midrule
GP bandit (ours)  & -110.995    & \textbf{7.655} & 5   & \textbf{10.99}                \\
grid-searched log & \textbf{-110.722}    & 8.389 & 100   & 177.01               \\ \bottomrule
\end{tabular}
\par\end{centering}
\end{table}

\bigskip

\begin{table}
\caption{Comparison between permutation invariant and non-permutation invariant
in MNIST dataset using S=$10$ (top) and S=$50$ (bottom). The best
scores are in bold. Given $T$ used epochs, the number of bandit
update and thus the number of sample for GP is $T/w$ where $w=10$
is the frequency update. The permutation invariant will be more favorable
when we have less samples for fitting the GP, as indicated in less
number of used epochs $T=1000,2000$. The performance is comparable
when we collect sufficiently large number of samples, e.g., when $T/w=1000$.} \label{table:ablation}
\begin{centering}

\begin{tabular}{cccccc}
\toprule
S=$10$ & Used Epoch $T$/ Bandit Iteration & $1000/100$ & $2000/200$ & $5000/500$ & $10000/1000$\tabularnewline
\midrule
\midrule
\multirow{2}{*}{d=$5$} & Perm Invariant & \textbf{$\boldsymbol{-91.488}$} & \textbf{$\boldsymbol{-90.129}$} & \textbf{$\boldsymbol{-89.130}$} & $-88.651$\tabularnewline
 & Non Perm Invariant & $-91.554$ & $-90.206$ & $-89.262$ & \textbf{$\boldsymbol{-88.552}$}\tabularnewline
\midrule
\multirow{2}{*}{d=$10$} & Perm Invariant & \textbf{$\boldsymbol{-91.430}$} & \textbf{$\boldsymbol{-90.219}$} & $-89.159$ & $-88.603$\tabularnewline
 & Non Perm Invariant & $-91.553$ & $-90.249$ & \textbf{$\boldsymbol{-89.110}$} & \textbf{$\boldsymbol{-88.466}$}\tabularnewline
\midrule
\multirow{2}{*}{d=$15$} & Perm Invariant & \textbf{$\boldsymbol{-91.386}$} & \textbf{$\boldsymbol{-90.059}$} & \textbf{$\boldsymbol{-88.957}$} & \textbf{$\boldsymbol{-88.504}$}\tabularnewline
 & Non Perm Invariant & $-91.550$ & $-90.224$ & $-89.215$ & $-88.564$\tabularnewline
\bottomrule
\end{tabular}
\par\end{centering}
\begin{centering}
\begin{tabular}{cccccc}
\toprule
S=$50$ & Used Epoch $T$/ Bandit Iteration & $1000/100$ & $2000/200$ & $5000/500$ & $10000/1000$\tabularnewline
\midrule
\midrule
\multirow{2}{*}{d=$5$} & Permutation Invariant & $\boldsymbol{-90.071}$ & \textbf{$\boldsymbol{-89.068}$} & \textbf{$\boldsymbol{-88.163}$} & $-87.979$\tabularnewline
 & Non Permutation Invariant & $-90.119$ & $-89.142$ & $-88.215$ & \textbf{$\boldsymbol{-87.860}$}\tabularnewline
\midrule
\multirow{2}{*}{d=$10$} & Perm Invariant & \textbf{$\boldsymbol{-90.125}$} & \textbf{$\boldsymbol{-89.115}$} & \textbf{$\boldsymbol{-88.187}$} & $-87.859$\tabularnewline
 & Non Permutation Invariant & $-90.212$ & $-89.225$ & $-88.231$ & \textbf{$\boldsymbol{-87.702}$}\tabularnewline
\midrule
\multirow{2}{*}{d=$15$} & Permutation Invariant & \textbf{$\boldsymbol{-90.029}$} & \textbf{$\boldsymbol{-89.082}$} & \textbf{$\boldsymbol{-88.102}$} & \textbf{$\boldsymbol{-87.579}$}\tabularnewline
 & Non Permutation Invariant & $-90.157$ & $-89.247$ & $-88.173$ & $-87.631$\tabularnewline
\end{tabular}
\par\end{centering}

\end{table}

\subsection{Training Curves}
\begin{figure*}[t]
\includegraphics[width=0.50\columnwidth]{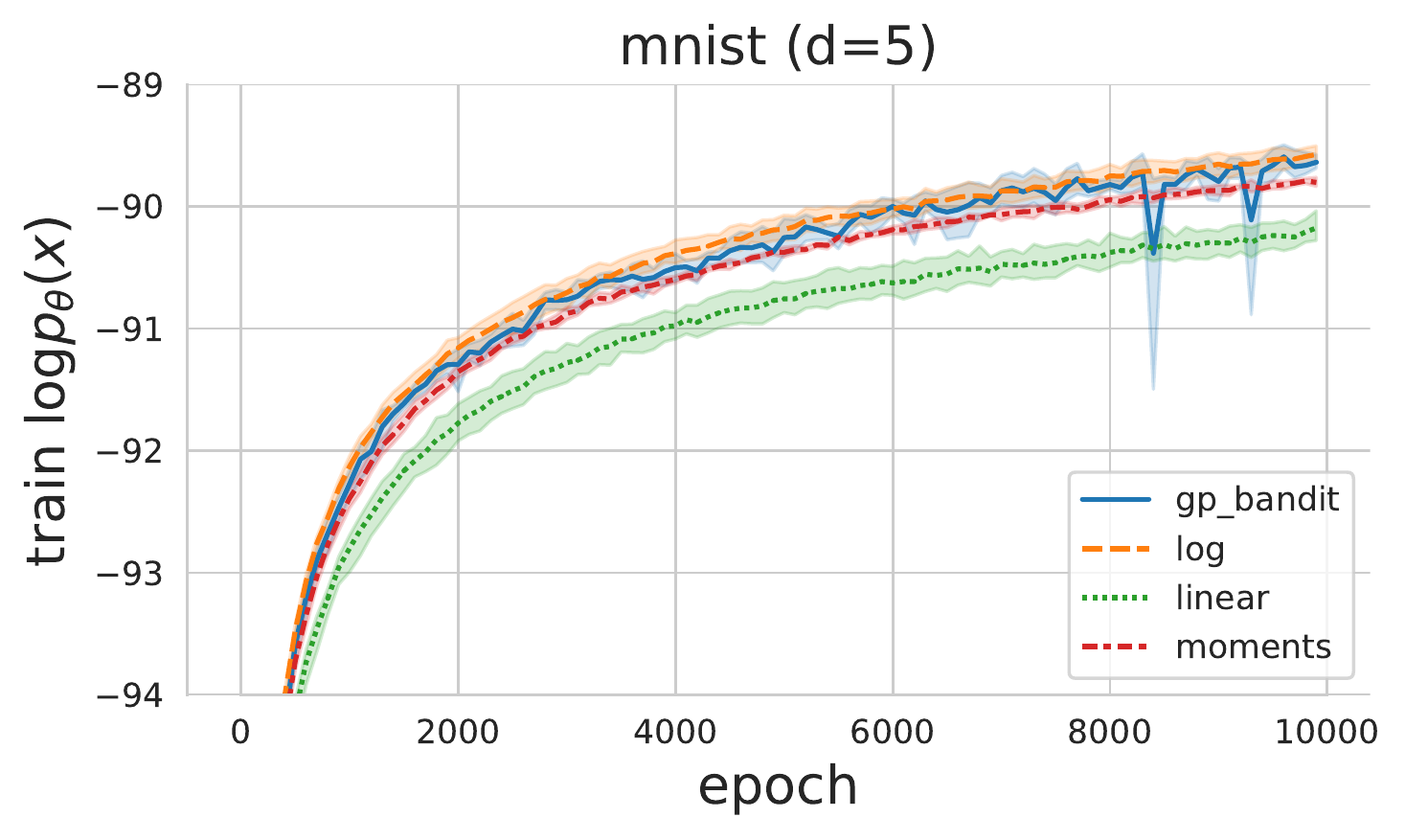}
\includegraphics[width=0.50\columnwidth]{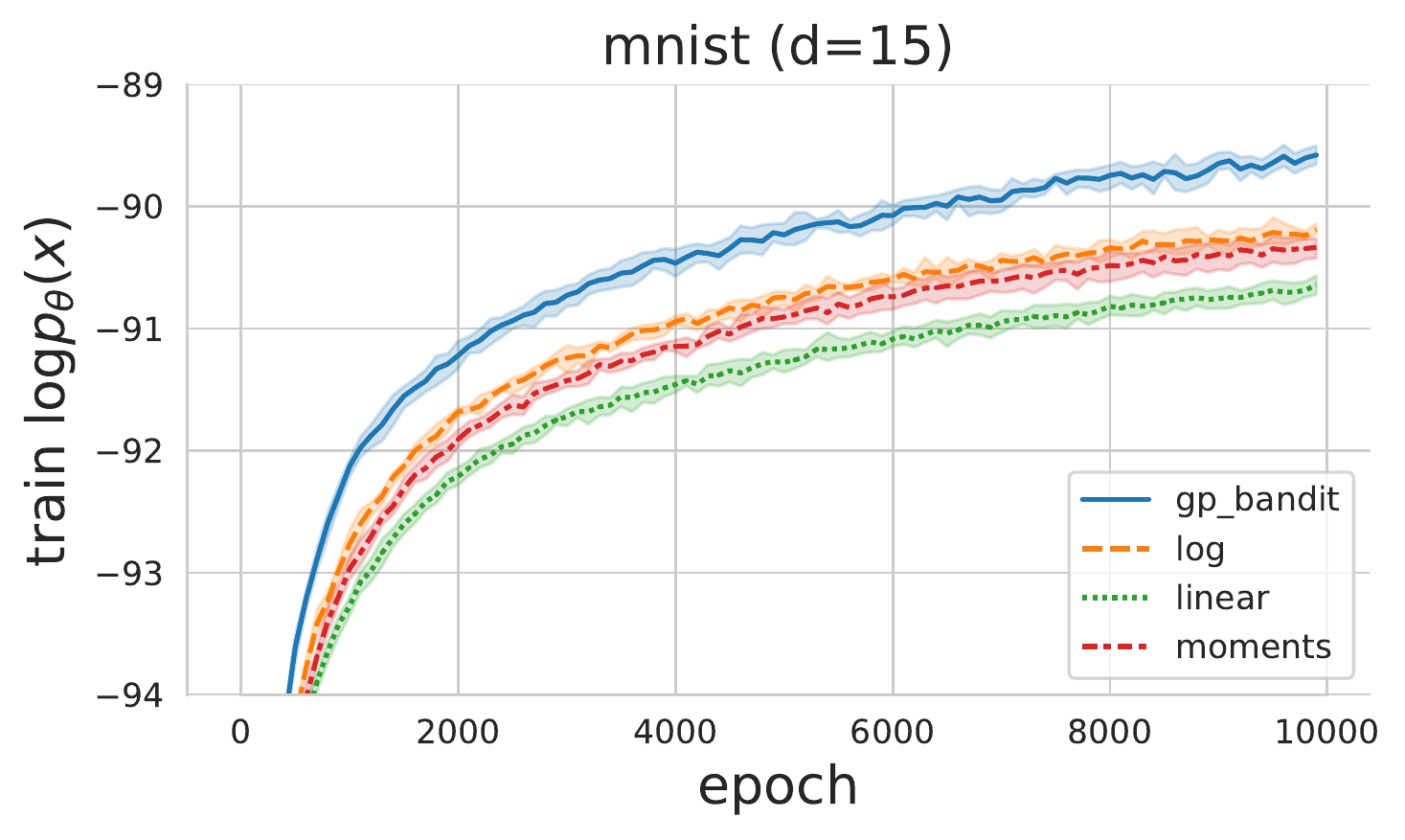}
\includegraphics[width=0.50\columnwidth]{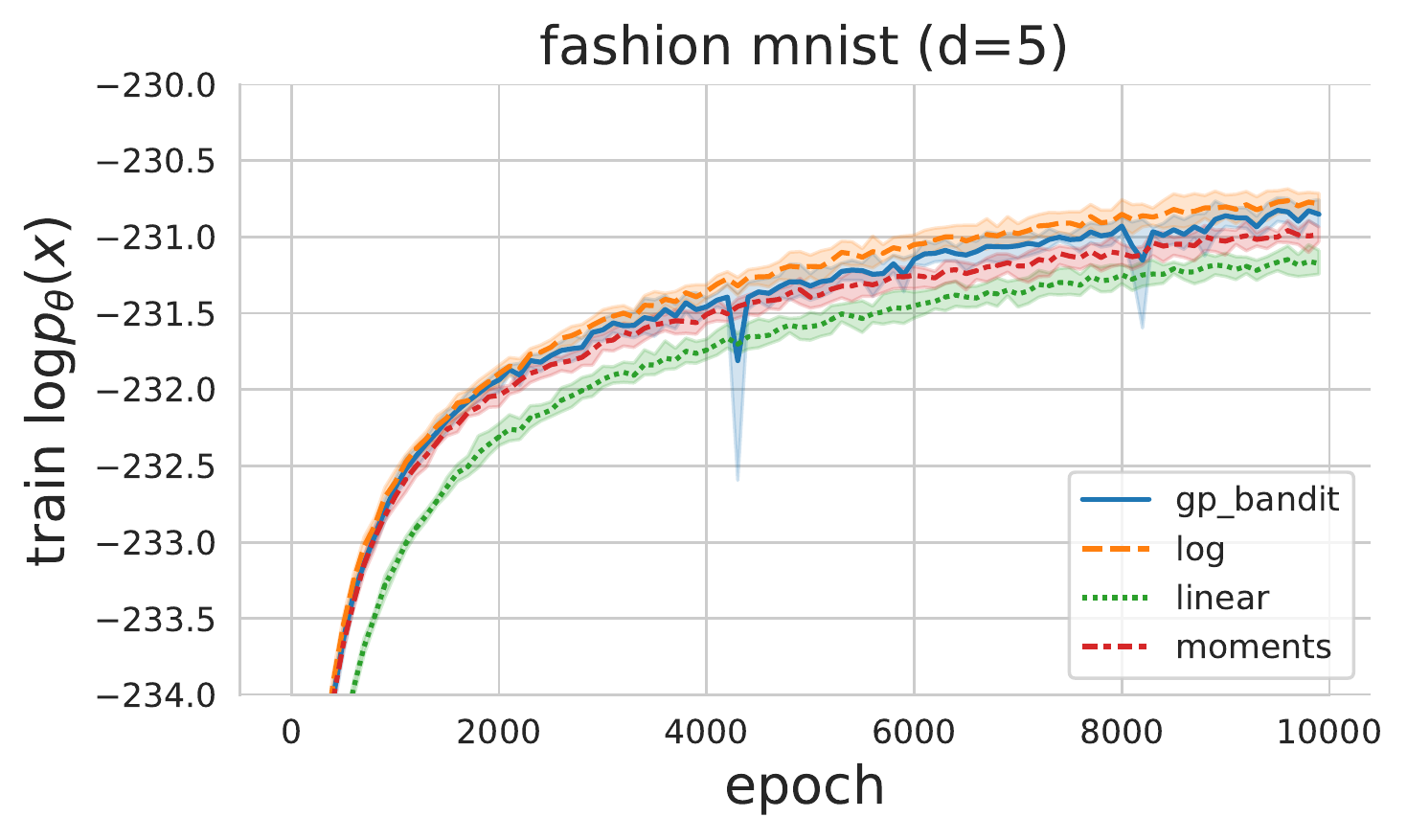}
\includegraphics[width=0.50\columnwidth]{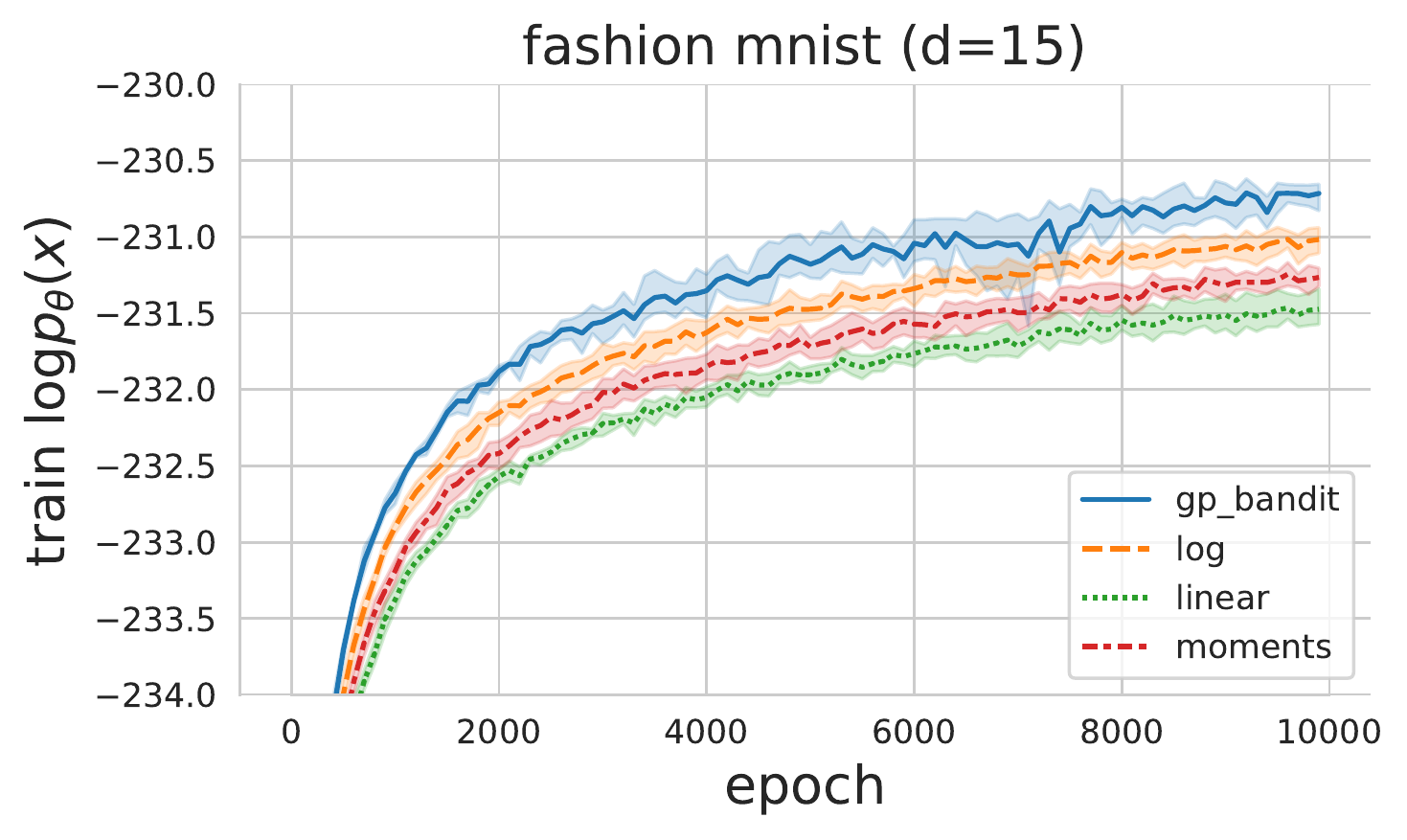}
\includegraphics[width=0.50\columnwidth]{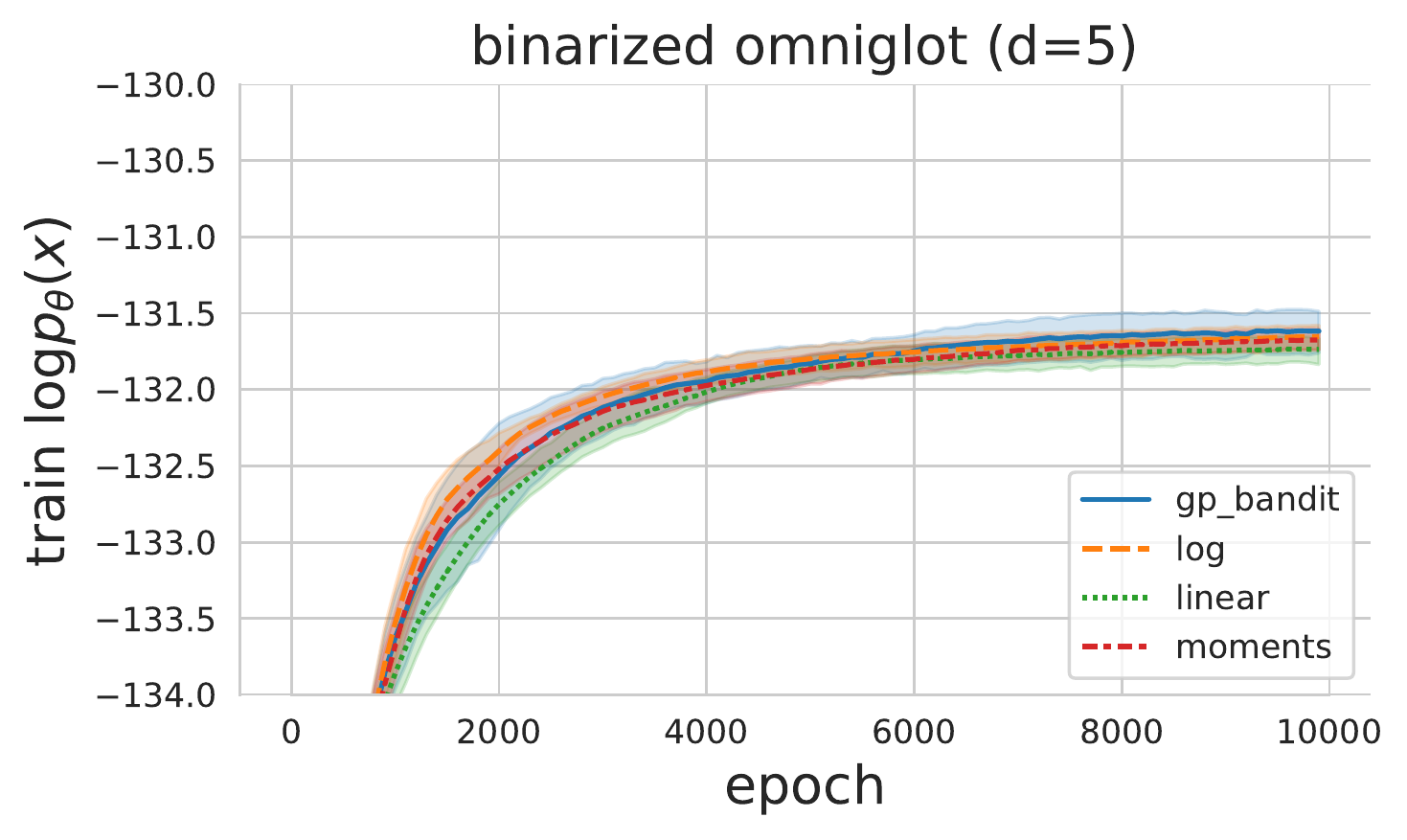}
\includegraphics[width=0.50\columnwidth]{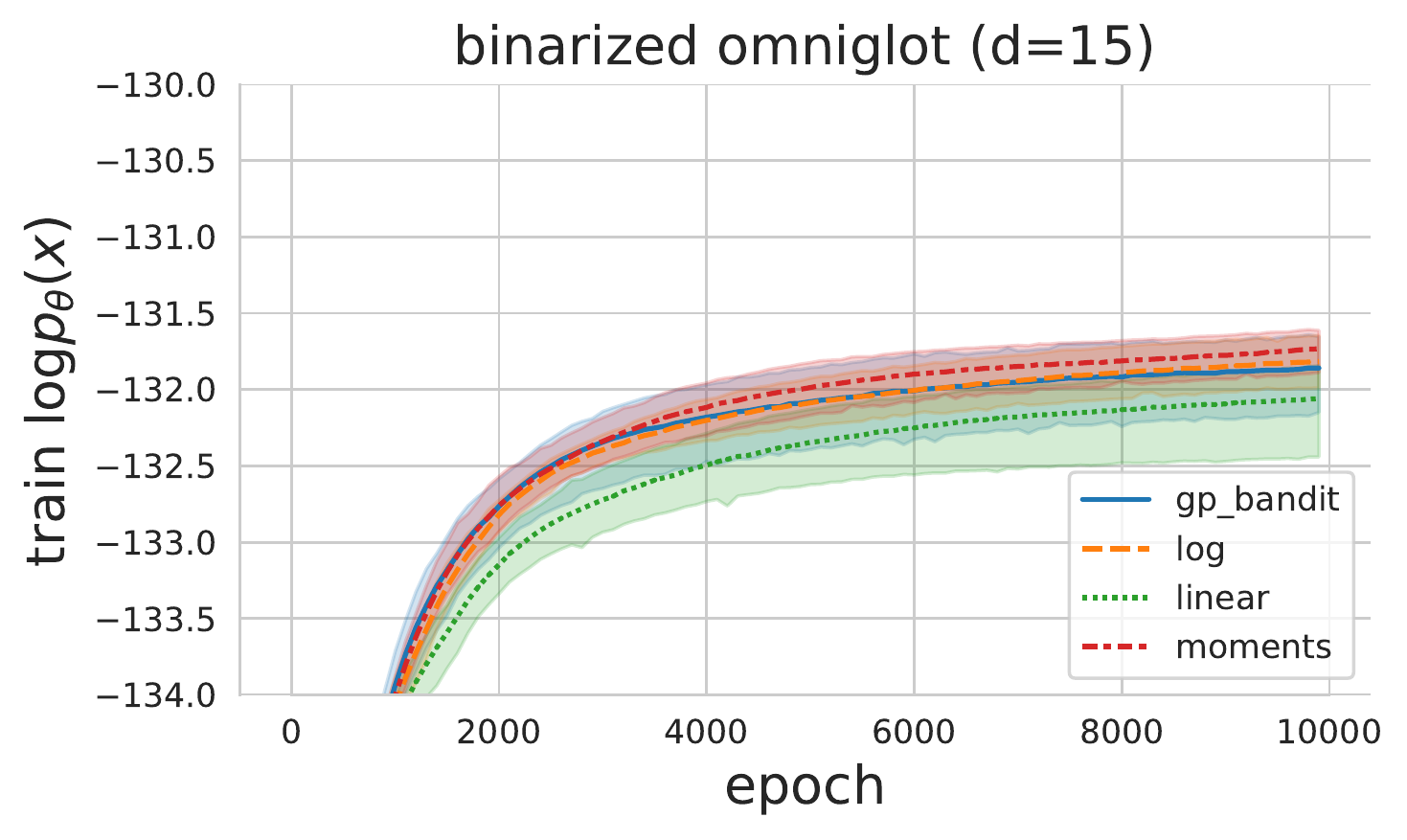}
\includegraphics[width=0.5\columnwidth]{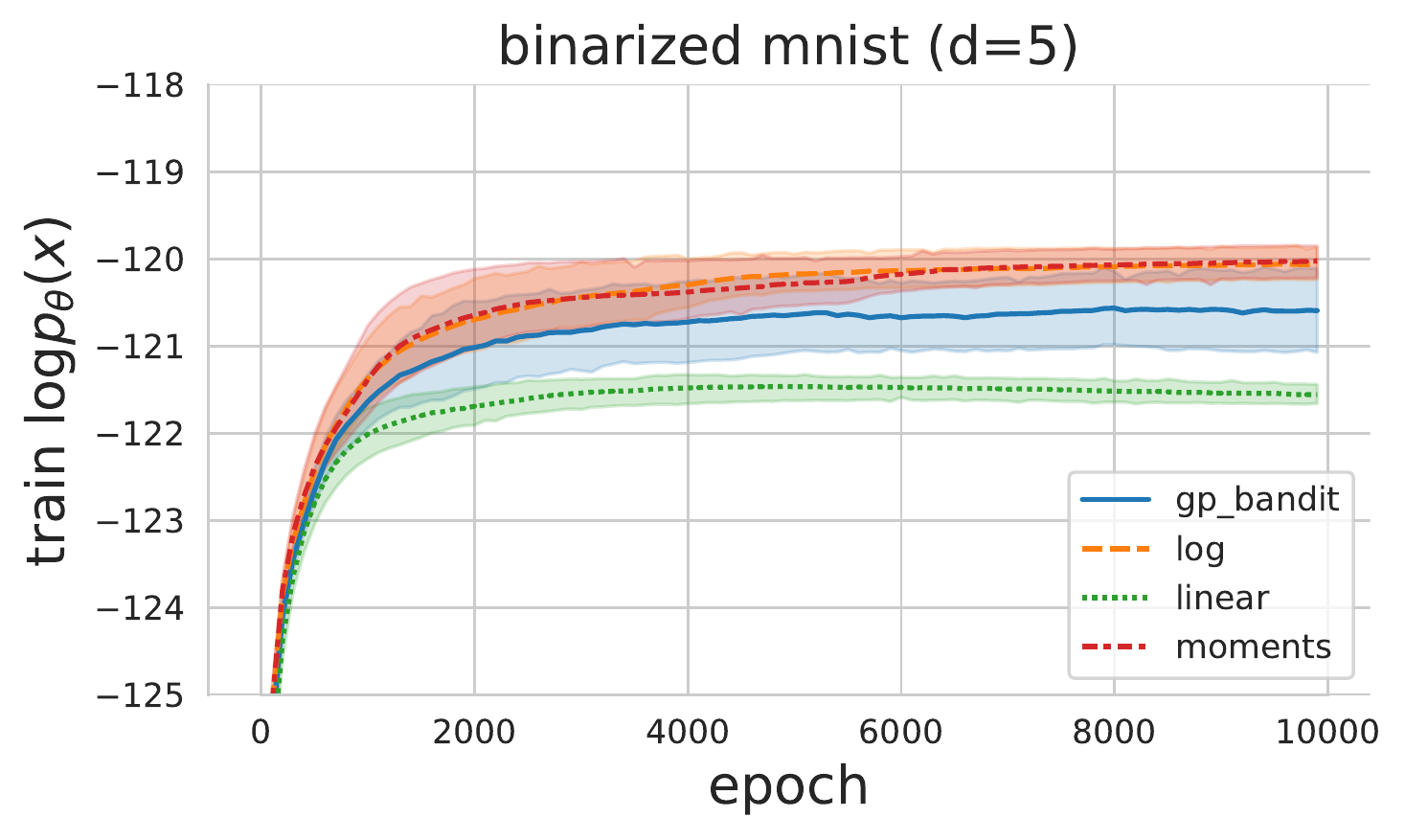}
\includegraphics[width=0.5\columnwidth]{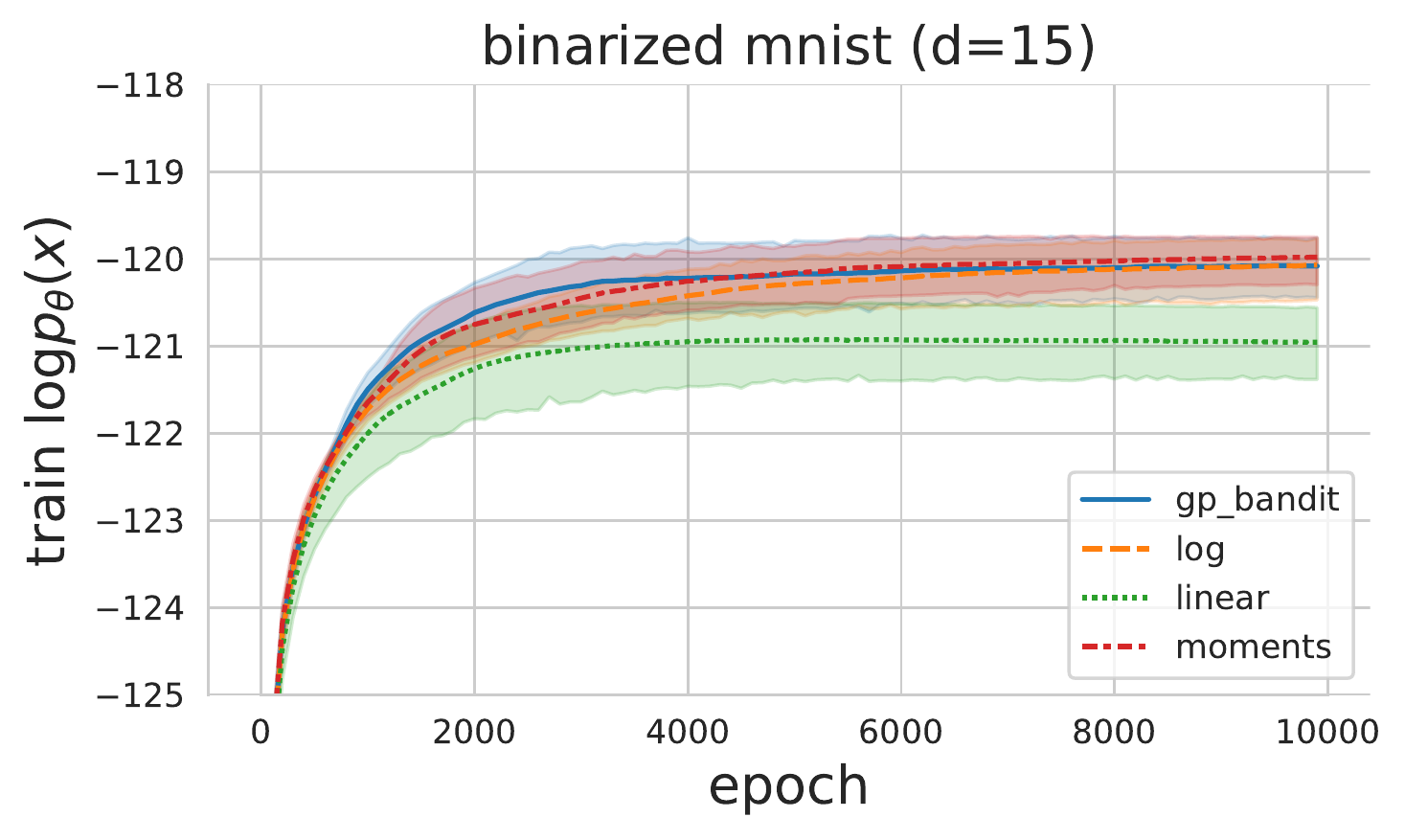}

\caption{We plot $\log p(\bx)$ on the training set throughout $S=10$, $d \in \{5,15\}$ for each dataset using the experimental setup described in \cref{app:ExperimentalSetup}. The final training log likelihoods are consistent with the test log likelihoods. Small drops in the GP-bandit training curves indicate the algorithm exploring the reward landscape (see Section \ref{sec:TVGP}).}
\label{fig:continuous_VAE_training_curves}
\end{figure*}

We show example training curves for $S=10, d \in \{5, 15\}$ obtained using the linear, log, moment, and GP-bandit schedules in Figure \ref{fig:continuous_VAE_training_curves}. We can see sudden drops in the GP-bandit training curves indicating our model is exploring alternate schedules during training (cf. Section \ref{sec:TVGP}).

\label{app:training_curves}
\end{document}